\documentclass{article}
\usepackage{iclr2027_conference,times}

\usepackage{amsmath,amsfonts,bm}

\def\eqref#1{equation~\ref{#1}}

\def\1{\bm{1}}

\DeclareMathAlphabet{\mathsfit}{\encodingdefault}{\sfdefault}{m}{sl}
\SetMathAlphabet{\mathsfit}{bold}{\encodingdefault}{\sfdefault}{bx}{n}

\newcommand{\E}{\mathbb{E}}

\newcommand{\R}{\mathbb{R}}

\usepackage{hyperref}
\usepackage{amsmath,amssymb,amsthm,mathtools}
\usepackage{booktabs}
\usepackage{graphicx}
\usepackage{xcolor}
\usepackage{colortbl}
\usepackage{multirow}
\usepackage{array}
\usepackage{arydshln}
\newcommand{\cdashlinelr}[1]{\cdashline{#1}}
\newcommand{\vertical}[1]{\rotatebox{90}{#1}}
\newcommand{\method}[1]{#1}
\newcommand{\model}[1]{\textbf{#1}}
\definecolor{mygreen}{RGB}{188,230,190}
\usepackage{algorithm}
\usepackage{algorithmic}
\usepackage{url}
\usepackage{wrapfig}

\newtheorem{proposition}{Proposition}
\newtheorem{lemma}{Lemma}
\newtheorem{assumption}{Assumption}
\newtheorem{remark}{Remark}

\newcommand{\tr}{\operatorname{tr}}
\newcommand{\rank}{\operatorname{rank}}
\newcommand{\vecop}{\operatorname{vec}}
\newcommand{\polar}{\operatorname{polar}}
\newcommand{\diag}{\operatorname{diag}}
\newcommand{\norm}[1]{\lVert #1 \rVert}
\newcommand{\Cb}{\bar{C}}
\newcommand{\Gb}{\bar{G}}
\newcommand{\unitop}{\operatorname{unit}}
\newcommand{\cond}{\operatorname{cond}}
\newcommand{\ftsv}{\textsc{F-TSV}}
\newcommand{\ftsvtwo}{\textsc{F-TSV}$^{2}$}

\title{\centering Task Singular Vectors go Functional:\\Model Merging in the Data Metric}

\author{
        Antonio Andrea Gargiulo$^1$ \And Alessandro Zirilli$^1$ \And Daniele Solombrino$^1$ \And
        \parbox{\linewidth}{%
        \vspace{-1.5em}
        \centering Emanuele Rodolà$^{1,2}$}\\
    \parbox{\linewidth}{%
        \small \centering
    $^1$Sapienza University of Rome, $^2$Paradigma \\
    \texttt{\small\{gargiulo, zirilli, solombrino, rodola\}@di.uniroma1.it}
    }
}

\iclrfinalcopy

\begin{document}
\maketitle
\lhead{Under review as a conference paper at ICLR 2027}  
\begin{abstract}
Spectral model merging compresses task vectors and reduces interference by orthogonalizing their retained directions. Existing methods make both decisions in the Euclidean geometry of weight space, however, directions that are orthogonal in parameter space can still overlap in the activations the model actually sees. We derive a \emph{joint least-squares} perspective that unifies several existing merging methods in a common objective, and use it to replace Euclidean geometry with a data-induced metric estimated from a few unlabeled samples. Functional TSV (\ftsv{}) utilizes this metric to choose each task’s low-rank subspace and to define cross-task orthogonality, while preserving the expert update within the chosen subspace. Our derivation predicts a square-root metric weighting, and validation sweeps across vision, GLUE, and generative merging select the same setting. Ablations reveal the main empirical effect: data-aware cross-task orthogonalization accounts for almost all of the gains in vision, whereas data-aware subspace selection contributes minimally outside GLUE. \ftsv{} is training-free, relies only on label-free metric, and produces a single checkpoint with no inference-time overhead, and it consistently improves over its Euclidean counterpart across all reported benchmarks.
\end{abstract}
\begin{wrapfigure}{r}{0.45\textwidth}
  \centering
    \vspace{-2.0em}
    \includegraphics[width=0.45\textwidth]{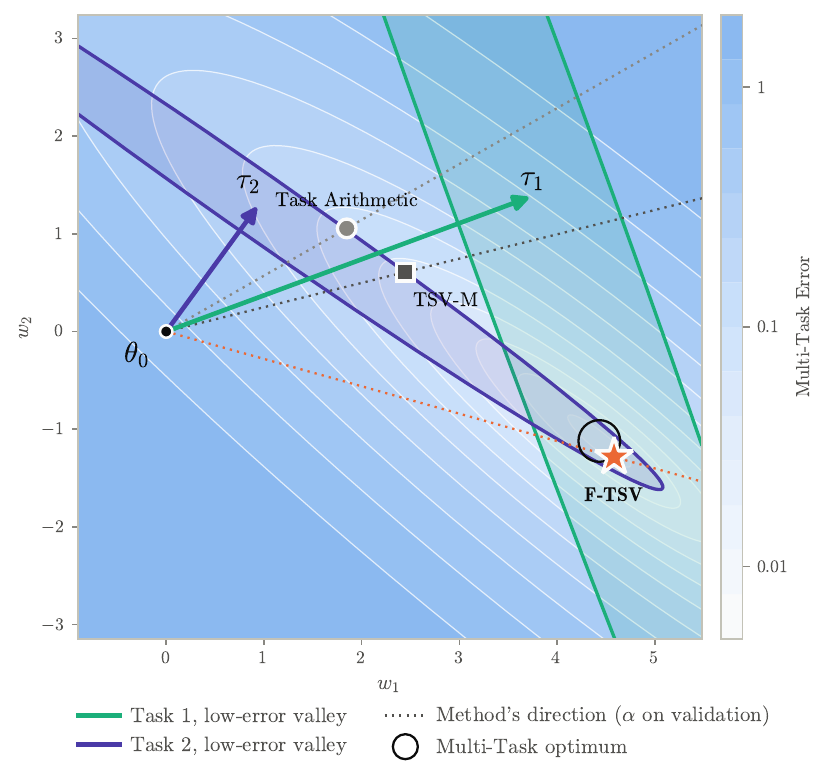}
    \vspace{-2.2em}
    \caption{Multi-Task loss landscape depicting final checkpoints after the corresponding model merging method in a 2D setting with 2 tasks. The arrows depict the two task vectors pointing to the low-error valley for their task. The dotted line is the direction each merging method chooses, and on that line, at the $\alpha$ value, is the final point.}
    \label{fig:toy}
    \vspace{-1.0em}
\end{wrapfigure}
\section{Introduction}
\label{sec:intro}
Finetuning a large pretrained model on different tasks produces a collection of specialist models. Model merging asks whether their task-specific knowledge can be combined into a single network that performs well across all tasks, without retraining or re-examining the labeled data \citep{wortsman2022soups,ilharco2023task}. The central difficulty is interference between tasks: as more experts are merged, updates useful for one task can conflict with the directions needed by another.
The central abstraction in model merging is the \emph{task vector} $\tau_t = \theta_t - \theta_0$, the representation of an expert as a displacement of task $t$'s finetuned weights $\theta_t$ from the shared pretrained model $\theta_0$. Merging is therefore not simply a matter of adding task vectors, but of deciding which parts of each update should survive and which retained directions should be allowed to coexist.
Recent spectral merging methods exploit the observation that these updates are approximately low-rank when viewed layer-by-layer \citep{hu2022lora}. TSV-M \citep{gargiulo2025tsv} retains a small number of singular directions from each task and then orthogonalizes the retained factors across tasks before recomposition to remove cross-task interference. Isotropic merging (Iso-C, \citet{marczak2025isotropic}) similarly operates on the spectrum of the merged update, flattening it. Despite their different constructions, these methods make the same two decisions: \emph{which directions of each update to retain}, and \emph{when two retained directions count as interfering}. In both cases, the decisions are based on the Euclidean inner product of weight space.

Euclidean geometry is convenient, but it does not necessarily reflect the function that the network actually computes. For a linear layer, the Euclidean top-\(k\) singular directions identify directions along which the weights activate most for an arbitrary unit input. The data, however, may rarely excite those directions. Likewise, Euclidean orthogonality removes correlation in parameter coordinates, but two parameter-space directions that satisfy \(v^\top v'=0\) can still produce strongly correlated changes on the inputs encountered by the model. The relevant overlap is instead determined by the data covariance \(C=\mathbb{E}[xx^\top]\). 
The issue is therefore not that spectral merging lacks a notion of geometry, but that it uses the parameters' geometry rather than the one induced by the data. Figure~\ref{fig:toy} clearly depicts this issue, showing how cross-task orthogonalization in the data metric helps in pointing towards a common minimum. 
We make this geometry explicit through a joint least-squares view of model merging. Starting from a second-order approximation of the task losses, we obtain a common objective in which the merged model is penalized for deviating from each expert according to a task-dependent curvature \(H_t\). This perspective places several existing merging methods in the same framework
under appropriate choices of curvature and constraints. 
We use this observation to replace the Euclidean geometry of spectral merging with a \emph{data-induced metric}. \textbf{Functional TSV} (\textbf{F-TSV}) uses the input metric to make the same two decisions as the other spectral merging methods \citep{gargiulo2025tsv, marczak2025isotropic, esm2026, an2025pave, cheng2025wudi}, in the geometry of the data: it selects the rank-$k$ subspace with the greatest data-weighted energy for task updates, and defines cross-task orthogonality using the pooled covariance of the task inputs.

We also introduce a whitening power $p$ that controls how the data metric enters the spectral construction. The Euclidean TSV-M geometry \citep{gargiulo2025tsv} corresponds to $p=0,$ while PAVE's ranking \citep{an2025pave} corresponds to $p=1$, and the joint least-squares derives the intermediate value $p=\tfrac12,$ 
This prediction is confirmed by held-out sweeps across vision, language, and generative merging, which select the same value.
The ablation also clarifies which part of the data-aware decisions matters: in vision, cross-task orthogonalization accounts for almost all of the improvement, while subspace selection contributes only marginally. This also extends to language and generative merging, where the relative contributions differ but the underlying geometric construction remains applicable. The distinction is informative because it separates two mechanisms that are often conflated in spectral merging: identifying useful directions within each task, and preventing the retained directions of different tasks from interfering. 
\ftsv{} requires only a small amount of unlabeled data to estimate these metrics. It is training-free, uses no task labels or task losses, and produces a single static checkpoint without inference-time routing or additional computation. The method, therefore, retains the computational structure of spectral merging while replacing its Euclidean notion of similarity with one grounded in the activations the network actually encounters.
To summarize, our contributions are fourfold:
\begin{itemize}\itemsep1pt
    \item We introduce \ftsv{}, a data-aware spectral merger (\S\ref{sec:ftsv}) that performs subspace selection and cross-task orthogonalization in a data-induced metric. 
    \item We derive a unified joint least-squares formulation that places several existing merging methods in a common objective and motivates the use of the data-induced metric (\S\ref{sec:jls}).
    \item We derive the whitening power $p=\tfrac12$, which interpolates between TSV-M's geometry and the PAVE's ranking; held-out validation confirms this value across all benchmarks (\S\ref{sec:power}).
    \item We introduce a label-free downstream-sensitivity extension based on $G=\E[J^{\top}J]$ from Gaussian vector, Jacobian probes (\S\ref{sec:ftsv2}), we show that the input metric can be estimated using only a small subset of unlabeled samples per task.
\end{itemize}

\section{Related work}
\label{sec:related}

Model merging began as weight averaging: model soups average finetuned models that share a pretrained initialization, without adding inference cost \citep{wortsman2022soups}, and task arithmetic represents each expert as a task vector $\tau_t$ and scales and sums them \citep{ilharco2023task}. Summed task vectors interfere as tasks accumulate, motivating methods that modify the updates at the level of parameters before aggregation. TIES trims small coordinates and resolves sign conflicts \citep{yadav2023ties}, DARE drops and rescales them \citep{yu2024dare}, and TALL masks localize each task's information in binary masks over the merged vector \citep{wang2024tall}.
A complementary view is to exploit the observation that finetuning updates are often closer to low-rank than to coordinate-sparse, particularly layer by layer \citep{hu2022lora,gargiulo2025tsv}, so spectral methods compress using an SVD rather than a mask. TSV-M keeps a few singular directions per task and orthogonalizes the stacked factors across tasks \citep{gargiulo2025tsv}. Iso-C flattens the spectrum of the summed update, and Iso-CTS adds task-specific subspaces to the shared one \citep{marczak2025isotropic}. 

A parallel line of work weights the merge by each task's curvature. Fisher merging uses the diagonal Fisher \citep{matena2022fisher}, RegMean matches each expert's layer outputs on its own inputs using Gram matrices \citep{jin2023regmean}, and MaTS solves the corresponding richer-curvature system by conjugate gradients \citep{tam2024mats}. \citet{daheim2024uncertainty} derives averaging, task arithmetic, and Fisher merging from a second-order expansion under different Hessian assumptions, while \citet{tam2024realistic} compares these methods under a common protocol, specifying data and compute requirements for each. 
Our joint least-squares formulation (\S\ref{sec:jls}) extends to spectral merging, borrowing the Kronecker-factored metric \citep{martens2015kfac} while keeping the per-task low-rank structure that exact curvature solvers abandon.
Closest to \ftsv{} are spectral methods influenced by activations. PAVE purifies each expert using its input covariance \citep{an2025pave}, it truncates the finetuned weight rather than the task vector, un-whitens the kept content by the inverse covariance, and has no cross-task orthogonalization. ESM restricts each update inside the principal subspace of the output changes it induces \citep{esm2026}, under an unconditioned metric, this corresponds to the exact rank-$k$ minimizer of our selection criterion at $p{=}\tfrac12$, but its subsequent orthogonalization remains Euclidean, making data-aware the step our ablation finds least useful (\S\ref{sec:mechanism}). Both make only the per-task step data-aware, cross-task orthogonalization in the data metric, the step \ftsv{} adds, is the one that brings the gain (see App.~\ref{app:blockortho} and~\ref{app:newbaselines}). WUDI- and ACE-Merging estimate input statistics directly from the task vectors \citep{cheng2025wudi,xu2026ace}, while other methods use activations to set merging coefficients \citep{nobari2025aim,yao2025agcm} or to route among task subspaces at inference time \citep{mass2025}. \ftsv{} uses activations only to define the inner product in which \emph{both} spectral decisions are made.

Our geometric interpretation is also related to functional maps in geometry processing. Functional maps represent correspondences between shapes as linear operators between function spaces, expressed in truncated eigenbases that are orthonormal under an area-weighted inner product \citep{ovsjanikov2012functional,ovsjanikov2017course}. Here, a task update is an operator between feature spaces, weight coordinates are the sampling, and $\langle\Delta,\Delta'\rangle_{C}=\E_x\langle\Delta x,\Delta' x\rangle$ is the area-weighted inner product in which both truncation and orthogonality are defined (Appendix~\ref{app:fmaps}). Spectral merging already truncates and orthogonalizes; what it lacks is this functional inner product induced by the data.

\section{Background}
\label{sec:notation}

Given a pretrained model $\theta_0$ and experts $\theta_1,\dots,\theta_T$ finetuned from it, \emph{merging} them produces a new set of weights $\theta = \theta_0 + \alpha \Delta^{\star}$, where $\Delta^{\star}$ aggregates the task vectors $\tau_t = \theta_t - \theta_0$, and $\alpha$ selected on a held-out validation set, as in Task Arithmetic, Tall-Mask, and TSV-M \citep{ilharco2023task, wang2024tall, gargiulo2025tsv}. In our setting, we are given an additional small \emph{unlabeled} subset of $N$ inputs from each task's input distribution $\mathcal D_t$. No labels are required, and therefore, neither task losses nor task-loss gradients are needed. Since aggregation is usually performed independently at each layer, for a weight matrix $W\in\R^{m\times n}$, we write $\Delta_t\in\R^{m\times n}$ for the corresponding layer block of the task vector $\tau_t$ (Table~\ref{tab:notation}). 
Spectral merging, as introduced by TSV-M \citep{gargiulo2025tsv}, treats each layer update as approximately low-rank and decomposes $\Delta_t = U_t S_t V_t^\top.$ Its construction proceeds in three steps: \emph{Compression} keeps the top $k = \max(1, \lfloor \min(m,n) / T \rfloor)$ singular triplets of the SVD from each task. This budget allocates the layer's spectrum equally among the $T$ experts, so that the retained blocks fill the spectrum exactly once. \emph{Stacking} collects the retained factors from all tasks, $\hat U=[U_1^{(k)}\! \cdots U_T^{(k)}],\! \quad \! \hat V=[V_1^{(k)}\! \cdots V_T^{(k)}]$, and stacks the retained singular values into $\hat s$. \emph{Interference removal} then replaces the stacked factors by their nearest orthonormal matrices, $\polar(A):=UV^{\top}$ for $A=USV^{\top}$, the resulting merged update is 
 \begin{equation}
     \Delta^{\star}_{\text{TSV-M}} = \polar(\hat U)\,\diag(\hat s)\,\polar(\hat V)^{\top}.
     \label{eq:tsvm}
 \end{equation}
These three steps implicitly make two geometric decisions. Compression determines \emph{which subspace of each task update is retained}, while interference removal determines \emph{when retained directions from different tasks are considered non-interfering}. Here, selecting the ``best'' rank-$k$ block and orthonormalizing are both defined in the Euclidean metric: two directions are non-interfering when $v^{\top}v'\! = \!0$. TSV-M, therefore, selects and orthogonalizes directions according to their geometry in \emph{parameter space}, independently of the inputs on which those directions will act.

\section{Method: Merging as joint least squares}
\label{sec:jls} 
The polar step of TSV-M drives $(V_t^{(k)})^{\top}V_{t'}^{(k)} \text{ to } 0$, for every pair of tasks $t\neq t'$, removing correlation in weight coordinates. We argue that two retained input directions $v,v'$ that are orthogonal in weight coordinates can still overlap \emph{under the data}: $\E_x[(v^{\top}x)((v')^{\top}x)]=v^{\top}\Cb v'$ with $\Cb$ the pooled input second moment, and the Euclidean metric has no access to that overlap. To resolve this inconsistency, we developed a framework that starts by expanding each task loss to a second-order approximation around its own expert, and regularizing the merged model towards the pretrained checkpoint, yielding the following \emph{joint least-squares} problem:
\begin{equation}
    \min_{\theta}\; \tfrac12 \sum_{t=1}^{T} (\theta - \theta_t)^{\top} H_t (\theta-\theta_t) \;+\; \mu\, \Omega (\theta-\theta_0),
    \label{eq:jls}
\end{equation}
where $H_t \succeq 0$ is a measure of curvature of the task $t$, and $\Omega$ a regularizer constraining the merged update. Writing $\theta = \theta_0 + \Delta$, the data term is $\sum_t(\Delta - \tau_t)^{\top} H_t (\Delta - \tau_t)$, and the merged \emph{update} is encouraged to remain close to every expert, each in its own data geometry, by the regularizer. 
The choices of $H_t$ and $\Omega$, place existing methods under a single framework. When fixing $H_t = I$ and $\Omega=0$ we obtain Task Arithmetic \citep{ilharco2023task}; $H_t = \diag(F_t)$, the diagonal Fisher, gives Fisher merging \citep{matena2022fisher}; $H_t = C_t \otimes I$, with $C_t$ the input second moment, we find RegMean \citep{jin2023regmean}; with $H_t$ set to the Kronecker-factored Fisher returns MaTS \citep{tam2024mats}; and $H_t = I$ with a per-task rank budget on $\Delta$ gives the geometry of TSV-M \citep{gargiulo2025tsv}, whose polar step exactly solves a factor-level problem, the nearest orthonormal factors (Proposition~\ref{prop:polar}).
Table~\ref{tab:ladder_theory} places each method on a ladder of approximations to \eqref{eq:jls}: the unconstrained methods solve it, the spectral ones use structured rules motivated by its geometry.

\textbf{Choosing the task geometry.}
Based on our theoretical framework, \ftsv{} is designed with the suitable geometry that merging requires. $H_t$ should measure how much a weight change alters the network's \emph{function}, not its coordinates. Consider a linear layer $y=Wx$ inside expert $t$, and let $f$ denote the network's final representation. Expanding the task loss to second order gives, for that layer, the Gauss--Newton curvature
\begin{equation}
  H_t = \E_{x\sim\mathcal{D}_t} \big[(xx^{\top}) \otimes (J^{\top} H_\ell J)\big] \;\approx\; C_t\otimes G_t, \qquad J := \partial f/\partial y,\ \ H_\ell:=\nabla_f^2\ell_t,
  \label{eq:kfac}
\end{equation}
so the data enters on one side through $x$, and the rest of the network enters on the other through the Jacobian $J$, which carries a perturbation of $y$ to the final representation. Here $\ell_t$ is the task loss on $f$, and we take $H_\ell\propto I$ (Assumption~\ref{as:iso}): the metric measures the squared change of $f$, a task-loss-agnostic and label-free choice.
$H_t$ is too large to store, so K-FAC \citep{martens2015kfac} assumes the input and output sides decouple, leaving one small matrix per side: $C_t=\E_{x\sim\mathcal{D}_t}[xx^{\top}]$ and $G_t=\E_{x\sim\mathcal{D}_t}[J^{\top}J]$.
The per-layer data term of \eqref{eq:jls} then becomes a \emph{two-sided} norm:
\begin{equation}
  (\Delta-\Delta_t)^{\top} H_t (\Delta-\Delta_t) = \big\|G_t^{1/2} (\Delta-\Delta_t) C_t^{1/2} \big\|_F^2.
  \label{eq:twosided}
\end{equation}
Each factor has a plain reading: multiplying on the right by $C^{1/2}$ converts ``change of weights'' into ``change of the layer's outputs on task data'',
and multiplying on the left by $G^{1/2}$ weights how much of that output change survives to the \emph{final representation}:
\begin{equation}
  \ftsv{} := \underbrace{\E_x\norm{\Delta x}^2}_{\text{output change}}=\norm{\Delta C^{1/2}}_F^2,
  \qquad
  \ftsv{}^2 := \underbrace{\E_x\norm{J\Delta x}^2}_{\text{representation change}}
  \approx\norm{G^{1/2}\Delta C^{1/2}}_F^2.
  \label{eq:functionalreading}
\end{equation}
The square-root exponents on $C$ and $G$ are the weighting prescribed by the least-squares geometry. In what follows, we generalize it to a free power $p$, with $p=\tfrac12$ as the derived default, and study its effect in \S\ref{sec:power}. The Euclidean merging is the case $p=0$, equivalently $C=G=I$: every parameter direction is treated equally, regardless of how the data excites or propagates it.
 
\textbf{Choosing the regularizer.}
The second ingredient is the constraint $\Omega$. Finetuning updates are sparse, not so much in coordinates (with few non-zero entries), the prior behind TIES and DARE \citep{yadav2023ties,yu2024dare}, as in the \emph{spectrum}: they concentrate in few singular directions \citep{gargiulo2025tsv}. We therefore take $\Omega$ to be TSV's rank budget: the merged update decomposes into one block per task, $\Delta=\sum_{t=1}^{T}\Delta_t^{(k)}$, with $\rank(\Delta_t^{(k)})\le k$, so that the $T$ blocks together fill the layer's spectrum exactly once. Each $\Delta_t^{(k)}$ holds what survives of expert $t$'s update $\Delta_t$, \emph{which} part survives is what \ftsv{} method decides (\S\ref{sec:ftsv}). 

\textbf{Why not solve \eqref{eq:jls} exactly.}
With $\Omega = 0$ and the one-sided metric, the exact minimizer of \eqref{eq:jls} is RegMean's $\Delta=(\sum_t\Delta_tC_t)(\sum_tC_t)^{-1}$ \citep{jin2023regmean}, which multiplies expert content by the full inverse of a \emph{sum of estimated} covariances, amplifying precisely the directions the estimates resolve worst. It also discards the per-task low-rank structure that denoises each expert update. We keep the constrained spectral form and use the data metric only where it is safe. 
The design principle is simple: the metric \emph{chooses} the subspaces and \emph{defines} orthogonality, restricting the expert update rather than reweighting it. Therefore, \ftsv{} should not be interpreted as the exact minimizer of \eqref{eq:jls}. Rather, the joint least-squares objective motivates the geometry, while the spectral budget, the non-interference constraint and the nearest-factor criterion (Proposition~\ref{prop:polar}) define a tractable merging rule, as in Algorithm~\ref{alg:ftsv}.

\subsection{Functional TSV}
\label{sec:ftsv}

\textbf{Step 1: metric-guided selection.} The Euclidean choice of $\Delta_t^{(k)}$ keeps the directions $v$ for which $\norm{\Delta_t v}$ is largest over unit vectors. The functional choice maximizes the same quantity, in the data geometry, over vectors of unit $C_t^{2p}$-norm, so the directions the data actually excites are easier to keep. We (i)~take the SVD of the whitened update $\Delta_t C_t^{\,p} = \tilde U \tilde S \tilde V^{\top}$ and retain its top-$k$ right singular vectors $\tilde V^{(k)}$; (ii)~we map them back to weight coordinates and orthonormalize them: $Q_t = \operatorname{QR} (C_t^{\,p} \tilde V^{(k)})$; and (iii)~keep the expert's action on that subspace \emph{undistorted}:
\begin{equation}
    \Delta_t^{(k)} = \Delta_t \, Q_t Q_t^{\top},
    \label{eq:select}
\end{equation}
We store the block in SVD form via $\Delta_t Q_t = U_t^{(k)} S_t^{(k)} (V_t^{(k)})^{\top}$, these are the factors that will be stacked and orthogonalized in Step~2, so the block's singular values are genuine singular values of $\Delta_t$ restricted to the subspace,
the exact rank-$k$ minimizer of the one-sided functional approximation error \eqref{eq:functionalreading} is 
the orthogonal projection $\tilde U^{(k)} (\tilde U^{(k)})^{\top}\Delta_t$ onto the leading left singular directions of the whitened update. The \eqref{eq:select} is equal to this exact block followed by a further projection onto $\mathrm{range}(Q_t)$: $\Delta_t Q_t Q_t^{\top} = (\tilde U^{(k)} (\tilde U^{(k)})^{\top} \Delta_t ) Q_t Q_t^{\top}$. 
Consequently, \eqref{eq:select} is weakly suboptimal under the isolated functional approximation criterion. By retaining the selected input directions themselves as the block's right factors, the next step can measure and remove \emph{cross-task} interference precisely within the subspace selected by the data-aware spectrum. The exact minimizer shares the column space but generally not the row space, so neither \eqref{eq:jls} nor Proposition~\ref{prop:exponent} decides between them. Gauged, \eqref{eq:select} is the better block under both orthogonalizations of the following Step~2, on average on vision and on GLUE (see Appendix~\ref{app:blockortho}).

\textbf{Step 2: interference removal.}
\label{sec:ortho}
Two selected directions $v,v'$ from different tasks can still respond to the same inputs, and their overlap is determined by the pooled input metric:
\begin{equation}
    \E_x\big[(v^{\top}x)((v')^{\top}x)\big]=v^{\top}\Cb v', \qquad \Cb=\tfrac1T\textstyle\sum_t\unitop(C_t),
    \label{eq:interference}
\end{equation}
with $\unitop(\cdot)$ the rescaling to unit mean eigenvalue prevents tasks with larger activation from dominating the shared geometry. The pooled metric $\Cb$ is then conditioned using \eqref{eq:conditioning}.
We redefine the notion of non-interference as $\Cb$-orthonormality and replace the stacked right factor by the nearest $\Cb^{\,2p}$-orthonormal matrix in the norm $\norm{(\cdot)\,\Cb^{\,p}}_F$, the metric analogue of the polar step\footnote{Writing $R'=R\,\Cb^{\,p}$ turns the constraint into $R'R'^{\top}=I$ and the objective into $\norm{R'-\hat V^{\top}\Cb^{\,p}}_F$, whose minimizer is $\polar(\hat V^{\top}\Cb^{\,p})$.}
\begin{equation}
    R=\polar\!\big(\hat V^{\top}\Cb^{\,p}\big)\,\Cb^{\,-p}, \qquad R\,\Cb^{\,2p}R^{\top}=I ,
    \label{eq:metricortho}
\end{equation}
which at the derived $p=\tfrac12$ is exactly the $\Cb$-orthonormality of \eqref{eq:interference}. The left factor keeps $L=\polar(\hat U)$ and the merged update is $\Delta^{\star} = L \diag(\hat s) R$, for an in-depth view see Algorithm~\ref{alg:ftsv}. So, \ftsv{} changes only the \emph{geometry} in which spectral merging operates: the data metric determines which directions are retained and which retained directions must be orthogonalized to reduce their interference, while the singular magnitudes $\hat s$ of the retained expert content are preserved.
\renewcommand{\arraystretch}{1.4}
\begin{table*}[t]
    \centering
    \vspace{-1.1em}
    \caption{\textbf{Vision benchmark results.} Average absolute accuracy, subscript in parentheses is the normalized average accuracy.}
    \resizebox{1\textwidth}{!} {
        \begin{tabular}{cc ccc ccc ccc}
            \toprule
            & \multirow{2}{*}{\textbf{Method}} & \multicolumn{3}{c}{\model{ViT-B-32}} & \multicolumn{3}{c}{\model{ViT-B-16}} & \multicolumn{3}{c}{\model{ViT-L-14}} \\
            \cmidrule(lr){3-5} \cmidrule(lr){6-8} \cmidrule(lr){9-11}
            & & 8 tasks & 14 tasks & 20 tasks & 8 tasks & 14 tasks & 20 tasks & 8 tasks & 14 tasks & 20 tasks \\
            \cmidrule(r){2-2} \cmidrule(lr){3-3} \cmidrule(lr){4-4} \cmidrule(lr){5-5} \cmidrule(lr){6-6} \cmidrule(lr){7-7} \cmidrule(lr){8-8} \cmidrule(lr){9-9} \cmidrule(lr){10-10} \cmidrule(lr){11-11} \rowcolor{gray!15}
            \cellcolor{white} & \multicolumn{1}{c}{\method{Zeroshot}} & $48.1_{(54.8)}$ & $57.0_{(64.5)}$ & $57.5_{(65.2)}$ & $55.4_{(60.7)}$ & $62.0_{(68.0)}$ & $62.5_{(68.3)}$ & $64.9_{(69.2)}$ & $69.1_{(73.8)}$ & $68.3_{(72.7)}$ \\
            \rowcolor{gray!15} \cellcolor{white} \multirow{-2}{*}{\small \vertical{Base}} & \multicolumn{1}{c}{\method{Finetuned}} & $90.3_{(100.0)}$ & $89.1_{(100.0)}$ & $89.5_{(100.0)}$ & $92.4_{(100.0)}$ & $91.4_{(100.0)}$ & $91.9_{(100.0)}$ & $94.2_{(100.0)}$ & $93.4_{(100.0)}$ & $94.0_{(100.0)}$ \\
            \cdashlinelr{1-11}
            \multirow{6}{*}{\small \vertical{Fixed}} & \multicolumn{1}{c}{\method{Weight Averaging}} & $67.1_{(74.6)}$ & $65.5_{(73.8)}$ & $64.4_{(72.6)}$ & $73.0_{(78.8)}$ & $70.8_{(77.3)}$ & $69.2_{(75.3)}$ & $80.5_{(85.3)}$ & $78.5_{(83.8)}$ & $76.1_{(80.9)}$ \\
            & \multicolumn{1}{c}{\method{Task Arithmetic}} & $70.4_{(77.9)}$ & $66.8_{(75.2)}$ & $64.0_{(71.9)}$ & $75.3_{(81.1)}$ & $72.0_{(78.6)}$ & $69.0_{(75.0)}$ & $84.4_{(89.3)}$ & $80.4_{(85.8)}$ & $76.8_{(81.7)}$ \\
            & \multicolumn{1}{c}{\method{TSV-M}} & $83.4_{(92.1)}$ & $79.1_{(88.6)}$ & $76.8_{(85.9)}$ & $85.8_{(92.5)}$ & $81.9_{(89.3)}$ & $80.0_{(86.9)}$ & $91.2_{(96.7)}$ & $88.8_{(94.9)}$ & $87.5_{(93.0)}$ \\
            & \multicolumn{1}{c}{\method{Iso-C}} & $83.7_{(92.5)}$ & $78.7_{(88.3)}$ & $74.9_{(83.9)}$ & $87.6_{(94.6)}$ & $81.7_{(89.1)}$ & $78.0_{(84.8)}$ & $92.4_{(97.9)}$ & $89.9_{(96.1)}$ & $86.8_{(92.3)}$ \\
            & \multicolumn{1}{c}{\method{Iso-CTS}} & $84.3_{(93.2)}$ & $80.7_{(90.5)}$ & $76.9_{(86.1)}$ & $88.7_{(95.8)}$ & $84.5_{(92.2)}$ & $81.0_{(88.0)}$ & $93.1_{(98.8)}$ & $90.5_{(96.8)}$ & $89.4_{(95.1)}$ \\
            \cdashlinelr{1-11}
            & \multicolumn{1}{c}{\method{ESM}} & $84.4_{(93.3)}$ & $80.1_{(89.9)}$ & $78.9_{(88.3)}$ & $87.6_{(94.6)}$ & $82.9_{(90.6)}$ & $81.3_{(88.3)}$ & $92.0_{(97.6)}$ & $89.2_{(95.4)}$ & $88.9_{(94.5)}$ \\
            & \multicolumn{1}{c}{\method{RegMean}} & $82.5_{(91.0)}$ & $77.2_{(86.5)}$ & $75.4_{(84.3)}$ & $85.3_{(92.0)}$ & $80.6_{(87.8)}$ & $78.0_{(84.7)}$ & $91.0_{(96.5)}$ & $88.1_{(94.1)}$ & $85.6_{(90.9)}$ \\
            & \multicolumn{1}{c}{\method{MaTS}} & $83.1_{(91.7)}$ & $77.8_{(87.1)}$ & $76.1_{(85.1)}$ & $86.2_{(93.0)}$ & $81.1_{(88.4)}$ & $78.9_{(85.7)}$ & $91.3_{(96.8)}$ & $88.4_{(94.5)}$ & $86.3_{(91.7)}$ \\
            \rowcolor{mygreen!50} \cellcolor{white} \multirow{-3}{*}{\small \vertical{Data}} & \multicolumn{1}{c}{\method{\textbf{Func. TSV}}} & $86.8_{(96.0)}$ & $83.6_{(93.7)}$ & $82.2_{(91.8)}$ & $\mathbf{89.1_{(96.4)}}$ & $85.6_{(93.6)}$ & $85.0_{(92.3)}$ & $93.2_{(98.9)}$ & $91.5_{(97.9)}$ & $\mathbf{91.9_{(97.7)}}$ \\
            \rowcolor{mygreen!50} \cellcolor{white}  & \multicolumn{1}{c}{\method{\textbf{Func. TSV}$\mathbf{^{2}}$}} & $\mathbf{87.0_{(96.2)}}$ & $\mathbf{83.8_{(93.9)}}$ & $\mathbf{82.4_{(92.0)}}$ & $88.7_{(96.0)}$ & $\mathbf{86.1_{(94.1)}}$ & $\mathbf{85.6_{(93.0)}}$ & $\mathbf{93.3_{(99.0)}}$ & $\mathbf{91.6_{(98.0)}}$ & $91.8_{(97.6)}$ \\
            \bottomrule
        \end{tabular}
    }
    \label{tab:vision}
\end{table*}

\subsection{Two-sided Functional TSV extension}
\label{sec:ftsv2}
The input metric asks which changes of a layer's output are large on task data, but it does not distinguish between changes that are subsequently preserved and those that are attenuated by the rest of the network. \ftsvtwo{} closes this gap by introducing the output-side sensitivity $G_t\!=\!\E[J^{\top}J]$.

For a Gaussian probe $v\!\sim\!\mathcal N(0,I)$ drawn in representation space, $\E_v[(J^{\top}v)(J^{\top}v)^{\top}]=J^{\top}J$, a Hutchinson-style identity \citep{hutchinson1990}, so the second moment of vector-Jacobian products estimates $G_t$ without materializing $J$. In practice, a single forward pass per batch stores the output $y$ of each linear layer (the \emph{sites}). Each of $n_p$ probes then costs a single backward pass, which produces $g=J^{\top}v$ at all sites at once. Accumulating $gg^{\top}$ estimate $G_t$ without labels or a task loss, making the estimator entirely label-free.
Selection now whitens on both sides, $G_t^{\,p}\Delta_tC_t^{\,p}=\tilde U\tilde S\tilde V^{\top}$, 
 with $Q_t$ and $\Delta_t^{(k)}$ constructed as in step~1. This changes the interpretation of the spectrum as: a direction scores highly only when the data excites it \emph{and} the layers above propagate its effect, no separate output-side truncation is needed (Lemma~\ref{lem:range}). The same geometry also defines interference on the output side, mirroring \eqref{eq:interference} with $\langle u,u'\rangle_{\Gb}=\E_x\langle J(x)u,J(x)u'\rangle$ and $\Gb=\tfrac1T\sum_t\unitop(G_t)$, we orthonormalize the stacked left factors in the pooled sensitivity metric, 
\begin{equation}
    L=\Gb^{\,-p}\polar\!\big(\Gb^{\,p}\hat U\big),\qquad L^{\top}\Gb^{\,2p}L=I,
    \label{eq:leftortho}
\end{equation}
which is the exact left-sided counterpart of \eqref{eq:metricortho}. Algorithm~\ref{alg:ftsv} completes the two-sided construction, setting $G_t = I$ removes the downstream sensitivity and recovers the one-sided \ftsv{} exactly.
The one-sided method is the main practical technique, \ftsvtwo{} explores whether downstream sensitivity further improves the same geometric framework.
\section{Experiments \& Results}
\label{sec:experiments}
The experiments test the geometric claim underlying \ftsv{}, each experiment below answers a specific question. 
We ask whether moving spectral merging from Euclidean weight space into the data metric improves the merge (\textbf{Q1}), 
which of the two data-aware decisions, subspace selection or cross-task orthogonalization, accounts for the improvement (\textbf{Q2}), where in the network the relevant geometry resides (\textbf{Q3}), whether the theoretically derived whitening power is recovered empirically (\textbf{Q4}), and whether the same geometry extends to generative merging (\textbf{Q5}).
\renewcommand{\arraystretch}{1.4}
\begin{table*}[t]
    \centering
    \vspace{-1.0em}
    \caption{\textbf{GLUE benchmark.} Headline score per task (Matthews for CoLA, Spearman for STS-B, accuracy elsewhere), subscript in parentheses indicates the normalized score. $\alpha$ represents the validation-selected coefficient.}
    \resizebox{1\textwidth}{!}{
        \begin{tabular}{cc cccccccc c c}
            \toprule
            & \multirow{2}{*}{\textbf{Method}} & \multicolumn{8}{c}{\model{RoBERTa-base} --- GLUE} & \multirow{2}{*}{\textbf{Avg.}} & \multirow{2}{*}{$\alpha$} \\
            \cmidrule(lr){3-10}
            & & CoLA & SST-2 & MRPC & STS-B & QQP & MNLI & QNLI & RTE & & \\
            \cmidrule(r){2-2} \cmidrule(lr){3-3} \cmidrule(lr){4-4} \cmidrule(lr){5-5} \cmidrule(lr){6-6} \cmidrule(lr){7-7} \cmidrule(lr){8-8} \cmidrule(lr){9-9} \cmidrule(lr){10-10} \cmidrule(lr){11-11} \cmidrule(lr){12-12} \rowcolor{gray!15}
            \cellcolor{white} & \multicolumn{1}{c}{\method{Pre-trained}} & $2.6_{(4.7)}$ & $50.9_{(53.8)}$ & $74.8_{(85.0)}$ & $0.0_{(0.0)}$ & $31.6_{(35.3)}$ & $32.4_{(37.0)}$ & $49.6_{(53.7)}$ & $47.3_{(71.2)}$ & $36.1_{(42.6)}$ & $-$ \\
            \rowcolor{gray!15} \cellcolor{white} \multirow{-2}{*}{\small \vertical{Base}} & \multicolumn{1}{c}{\method{Finetuned}} & $55.1$ & $94.7$ & $88.0$ & $87.7$ & $89.5$ & $87.5$ & $92.4$ & $66.4$ & $82.7$ & $-$ \\
            \cdashlinelr{1-12}
            \multirow{5}{*}{\small \vertical{Fixed}} & \multicolumn{1}{c}{\method{Model Soup}} & $2.6_{(4.7)}$ & $55.3_{(58.4)}$ & $75.5_{(85.8)}$ & $42.0_{(47.9)}$ & $40.0_{(44.7)}$ & $41.8_{(47.7)}$ & $59.8_{(64.7)}$ & $47.3_{(71.2)}$ & $45.5_{(53.1)}$ & $-$ \\
            & \multicolumn{1}{c}{\method{Task Arithmetic}} & $2.6_{(4.7)}$ & $85.6_{(90.3)}$ & $78.9_{(89.7)}$ & $26.2_{(29.8)}$ & $72.9_{(81.5)}$ & $62.3_{(71.2)}$ & $76.2_{(82.5)}$ & $52.3_{(78.8)}$ & $57.1_{(66.1)}$ & $0.4$ \\
            & \multicolumn{1}{c}{\method{TSV-M}} & $21.5_{(39.0)}$ & $92.8_{(97.9)}$ & $78.1_{(88.8)}$ & $43.9_{(50.1)}$ & $82.8_{(92.5)}$ & $76.1_{(87.0)}$ & $83.6_{(90.4)}$ & $52.0_{(78.3)}$ & $66.4_{(78.0)}$ & $1.4$ \\
            & \multicolumn{1}{c}{\method{Iso-C}} & $26.8_{(48.7)}$ & $89.6_{(94.6)}$ & $78.2_{(88.9)}$ & $58.2_{(66.4)}$ & $84.4_{(94.4)}$ & $76.2_{(87.1)}$ & $83.5_{(90.3)}$ & $\mathbf{61.7_{(92.9)}}$ & $69.8_{(82.9)}$ & $3.1$ \\
            & \multicolumn{1}{c}{\method{Iso-CTS}} & $40.5_{(73.5)}$ & $84.4_{(89.1)}$ & $78.8_{(89.6)}$ & $49.0_{(55.8)}$ & $82.6_{(92.3)}$ & $67.7_{(77.4)}$ & $78.8_{(85.2)}$ & $57.4_{(86.4)}$ & $67.4_{(81.2)}$ & $3.4$ \\
            \cdashlinelr{1-12}
            & \multicolumn{1}{c}{\method{ESM}} & $34.1_{(61.8)}$ & $70.8_{(74.7)}$ & $75.6_{(85.9)}$ & $34.9_{(39.8)}$ & $63.6_{(71.1)}$ & $63.1_{(72.2)}$ & $51.7_{(55.9)}$ & $47.3_{(71.2)}$ & $55.1_{(66.6)}$ & $0.5$ \\
            & \multicolumn{1}{c}{\method{RegMean}} & $44.7_{(81.1)}$ & $\mathbf{93.9_{(99.2)}}$ & $80.7_{(91.7)}$ & $62.1_{(70.8)}$ & $82.9_{(92.7)}$ & $79.7_{(91.1)}$ & $87.9_{(95.1)}$ & $41.2_{(62.0)}$ & $71.6_{(85.5)}$ & $1.4$ \\
            & \multicolumn{1}{c}{\method{MaTS}} & $43.5_{(79.0)}$ & $93.5_{(98.7)}$ & $81.8_{(92.9)}$ & $60.4_{(68.9)}$ & $83.3_{(93.1)}$ & $80.0_{(91.5)}$ & $\mathbf{88.0_{(95.1)}}$ & $43.0_{(64.7)}$ & $71.7_{(85.5)}$ & $1.3$ \\
            \rowcolor{mygreen!50}
            \cellcolor{white} \multirow{-3}{*}{\small \vertical{Data}} & \multicolumn{1}{c}{\method{\textbf{Func. TSV}}} & $45.4_{(82.4)}$ & $92.5_{(97.7)}$ & $\mathbf{82.1_{(93.4)}}$ & $58.6_{(66.8)}$ & $\mathbf{84.5_{(94.4)}}$ & $\mathbf{81.3_{(92.9)}}$ & $85.7_{(92.7)}$ & $54.9_{(82.6)}$ & $73.1_{(87.9)}$ & $3.0$ \\
            \rowcolor{mygreen!50}
            \cellcolor{white} & \multicolumn{1}{c}{\method{\textbf{Func. TSV$^{2}$}}} & $\mathbf{46.1_{(83.7)}}$ & $92.3_{(97.5)}$ & $80.0_{(90.9)}$ & $\mathbf{63.5_{(72.4)}}$ & $84.1_{(94.0)}$ & $80.0_{(91.4)}$ & $86.3_{(93.4)}$ & $58.8_{(88.6)}$ & $\mathbf{73.9_{(89.0)}}$ & $3.2$ \\
            \bottomrule
        \end{tabular}
    }
    \label{tab:glue}
    \vspace{-1.1em}
\end{table*}
\paragraph{Benchmarks setup.}
\label{sec:setup}

We evaluate the same geometric construction across three settings. For \emph{vision:} we used CLIP encoders (ViT- B/32, B/16, L/14) merged over the 8-task suite \citep{ilharco2023task} and its 14- and 20-task extensions \citep{wang2024tall}, evaluated with frozen per-task heads. For \emph{language} we opted for RoBERTa-base experts on 8 GLUE tasks \citep{wang2018glue}, following the benchmark's task-specific metrics: Matthews for CoLA, Spearman for STS-B, and accuracy elsewhere. For \emph{generative} we used five Llama-3.2-3B MergeBench experts \citep{mergebench2025} and report the benchmark's five-pillar score.

\textbf{Protocol.} For every method with a merge coefficient, we select $\alpha$ on a held-out validation split. On vision and GLUE, the validation split is a seeded $10\%$ subset of each task's training data. On MergeBench, where evaluating each candidate $\alpha$ is substantially more expensive, we select $\alpha$ based on the teacher-forced validation likelihood. The reported results are from the test split evaluated once at the selected validation $\alpha$. We report the absolute and the \emph{normalized} score, \emph{i.e.}\ the merged score divided by the corresponding expert and averaged across tasks. 

\paragraph{Main results (Q1).}
\label{sec:mainresults}
\begin{wrapfigure}{r}{0.5\textwidth}
    \centering
    \vspace{-1.0em}
    \includegraphics[width=0.5\textwidth]{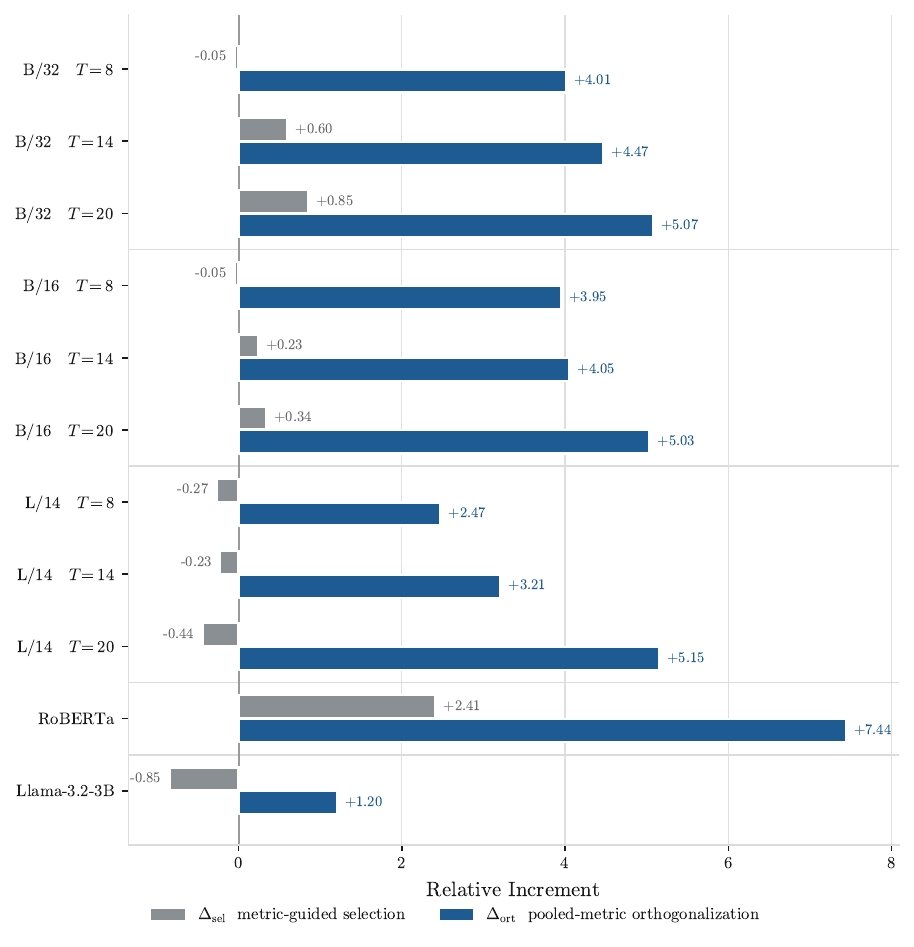}
    \vspace{-2.0em}
    \caption{Comparison between metric-guided selection and the pooled-metric orthogonalization. $\Delta_{\text{sel}}$ and $\Delta_{\text{ort}}$ are the two increments in normalized points w.r.t. Euclidean polar (TSV).}
    \vspace{-0.8em}
    \label{fig:mechanism}
\end{wrapfigure}
We first ask whether replacing Euclidean spectral geometry with the data metric improves the merged model.
On \emph{vision} (Table~\ref{tab:vision}), the \ftsv{} family obtains the highest score among the evaluated methods in all nine settings, improving normalized accuracy over the best merging methods. 
The margin over TSV-M, the $p=0$ Euclidean case, increases with task count across all backbones.
This scaling is consistent with the role of cross-task interference: as more tasks compete for the same spectral budget, the value of defining the cross-task decorrelation under the data geometry increases. Figure~\ref{fig:mechanism} shows that this growing margin is associated with cross-task decorrelation rather than with changing which subspace each task retains.
On \emph{Language} (Table~\ref{tab:glue}), \ftsv{} raises the average score to $73.1$ ($87.9$ normalized), while \ftsvtwo{} reaches $73.9$ ($89.0$ normalized). The strongest baselines, RegMean \citep{jin2023regmean} and MaTS \citep{tam2024mats}, reach $71.6$ and $71.7$ ($85.5$ normalized), the two curvature-aware families trade places by task: RegMean is ahead on STS-B and QNLI, while \ftsv{} on MNLI and MRPC. Notably, unlike the vision results, RegMean, a data-driven method, is a weaker baseline on vision (Table~\ref{tab:vision}) than on GLUE; by contrast, our method is more balanced between the two benchmarks.
    
\paragraph{Ablation (Q2).}
\label{sec:mechanism}
The central question here is whether the improvement comes from choosing better subspaces for each task, from preventing interference between the retained subspaces, or from both. Figure~\ref{fig:mechanism} separates the two data-aware decisions of \S\ref{sec:ftsv} on every benchmark. On vision, the contrast is sharp: averaged over the nine cells, metric-guided selection is worth $+0.11$ normalized points and the pooled-metric orthogonalization $+4.16$. Selection is small and directionless, spanning $-0.44$ to $+0.85$ and with five \emph{negative} outcomes, whereas orthogonalization improves every setting, and increases with task count.
What scaling the number of tasks makes more valuable is the \emph{decorrelation} of the retained subspaces under the data metric, rather than the per-task choice of which subspace to retain.
The effect persists across modalities, although the margin does not. On MergeBench, orthogonalization remains the positive contribution ($+1.20$ on average), while selection does not ($-0.85$). GLUE is the setting where selection contributes substantially: $+2.41$ of the $9.85$ normalized improvement points. Across the three modalities, the orthogonalization is consistently positive, whereas selection is positive only on GLUE. This bears directly on per-task purification methods such as PAVE \citep{an2025pave}, whose ranking rule is close to our metric-guided selection and which have no cross-task stage (see Appendix~\ref{app:newbaselines}). In our experiments, cross-task orthogonalization accounts for substantially more of the observed improvement than metric-guided selection.
The complete ablation below crosses the two steps with the same rank budget, statistics and protocol. Orthogonalization alone, on the Euclidean top-$k$ blocks, improves TSV-M by $+4.19$ on vision and $+7.36$ on GLUE, and data-aware selection adds $+0.08$ and $+2.49$ on top of it. On the generative benchmark the two rows select different coefficients ($\alpha=1.2$ and $1.8$; at $\alpha=1.2$, \ftsv{} scores $36.03$), so its $+3.08$ is not a gain of selection alone.
\begin{center}
    \small
    \begin{tabular}{@{}ll ccc@{}}
        \toprule
        Selection & Orthogonalization & Vision (mean of nine) & GLUE & MergeBench \\
        \midrule
        Euclidean & Euclidean (TSV-M) & 91.09 & 78.01 & 39.45 \\
        Data & Euclidean & 91.20 & 80.42 & 38.60 \\
        Euclidean & Data & 95.28 & 85.37 & 36.72 \\
        Data & Data (\ftsv{}) & \textbf{95.36} & \textbf{87.86} & \textbf{39.80} \\
        \bottomrule
    \end{tabular}
\end{center}
Switching a metric off reproduces the parent method as Proposition~\ref{prop:nesting} prescribes (Appendix~\ref{app:nesting}).
\paragraph{Where the geometry lives (Q3).}
\label{sec:sites}
\renewcommand{\arraystretch}{1.4}
\begin{table}[t]
    \centering
    \vspace{-1.1em}
    \caption{\textbf{Site attribution on vision:} the MLP carries the input geometry, and the output metric is near-flat. Which weight matrices keep their measured metric, each row is that metric's marginal value at those sites, and each \method{none} row reproduces its parent. Absolute score with normalized in subscript. \emph{Gain} is the normalized-score change against the \method{parent}.}
        \begin{tabular}{cl cc @{\hskip 1.6em} cl cc}
            \toprule
            & \multirow{2}{*}{\textbf{Sites kept}} & \multicolumn{2}{c}{\model{Input metric $C$}} & & \multirow{2}{*}{\textbf{Sites kept}} & \multicolumn{2}{c}{\model{Output metric $G$}} \\
            & & \multicolumn{2}{c}{\small ViT-B/32, 8 tasks} & & & \multicolumn{2}{c}{\small ViT-B/32, 8 tasks} \\
            \cmidrule(lr){2-2}\cmidrule(lr){3-4} \cmidrule(lr){6-6}\cmidrule(lr){7-8}
            & & Score & Gain & & & Score & Gain \\
            \cmidrule(r){1-4} \cmidrule(l){6-8}
            \rowcolor{gray!15}
            \cellcolor{white} & \method{none ($\equiv$\,TSV-M)} & $83.41_{(92.05)}$ & --- & \cellcolor{white} & \method{none ($\equiv$\,\ftsv{})} & $86.81_{(96.02)}$ & --- \\
            & \method{q/k/v} & $83.62_{(92.27)}$ & $+0.21$ &  & \method{q/k/v} & $86.51_{(95.65)}$ & $-0.37$ \\
            & \method{all attention} & $83.90_{(92.61)}$ & $+0.55$ &  & \method{all attention} & $86.66_{(95.84)}$ & $-0.18$ \\
            & \method{non-residual} & $84.72_{(93.56)}$ & $+1.51$ &  & \method{non-residual} & $86.82_{(96.02)}$ & $+0.00$ \\
            & \method{residual} & $85.78_{(94.74)}$ & $+2.68$ &  & \method{residual} & $86.68_{(95.86)}$ & $-0.15$ \\
            & \method{q/k/v $+$ residual} & $85.85_{(94.79)}$ & $+2.73$ &  & \method{q/k/v $+$ residual} & $86.67_{(95.85)}$ & $-0.17$ \\
            \rowcolor{mygreen!50}
            \cellcolor{white} & \method{MLP linears} & $86.25_{(95.29)}$ & $+3.24$ & \cellcolor{white} & \method{MLP linears} & $\mathbf{86.99_{(96.25)}}$ & $+0.23$ \\
            \rowcolor{mygreen!50}
            \cellcolor{white} & \method{MLP $+$ residual} & $86.54_{(95.65)}$ & $+3.59$ & \cellcolor{white} & \method{MLP $+$ residual} & $86.83_{(96.03)}$ & $+0.02$ \\
            \rowcolor{mygreen!50}
            \cellcolor{white} & \method{all sites} & $\mathbf{86.81_{(96.02)}}$ & $+3.96$ & \cellcolor{white} & \method{all sites} & $86.96_{(96.22)}$ & $+0.20$ \\
            \bottomrule
        \end{tabular}
    \label{tab:sites}
    \vspace{-1.1em}
\end{table}
We ask \emph{where in the network} the geometry resides, by limiting the metric to a subset of sites.
Table~\ref{tab:sites} and Table~\ref{tab:sites_glue} show \emph{which} matrices preserve their measured metric, for vision and GLUE. On vision, the input geometry is concentrated in the MLP and in the sites contributing to the residual stream. Keeping $C$ only at the attention projections recovers $0.55$ of the available $3.96$ points, while the MLP alone recovers $3.24$, and MLP plus residual $3.59$, with no combination reaching the full gain. The output metric is comparatively flat: peaking at $+0.23$ with the MLP, while several subsets perform slightly \emph{below} the one-sided \ftsv{} method.
On GLUE, the input metric tells a different story: the stronger proper subset is attention q/k/v plus residual, recovering $+7.13$ of the available $9.85$. Here, unlike in vision, the MLP alone is not the dominant site, contributing $+4.57$. Adding $G$ produces a minor change: at all sites, it recovers $+1.12$ over the one-sided construction, most of it from the MLP. This ablation suggests that the post-normalization residual streams are anisotropic and task-discriminative across modalities, with modality-dependent contributions from MLP or attention. In practice, collecting statistics at MLP sites alone recovers $82\%$ of the vision gain at lower cost, while on GLUE the best subset recovers $72\%$.
\paragraph{The whitening power(Q4).}
\label{sec:power}
The derivation of \S\ref{sec:jls} predicts how strongly the data metric should enter the spectral construction. The whitening power $p$ in $C^{\,p}$ should be $\tfrac12$, the square-root geometry induced by the least-squares objective of \eqref{eq:functionalreading}, rather than the $p{=}1$ ranking of PAVE \citep{an2025pave} or the $p{=}0$ of TSV \citep{gargiulo2025tsv}. This is the unique power for which the metric-polar constraint also coincides exactly with the proposed functional orthogonality criterion. Sweeping $p$ on all three modalities gives a grid showing the derived value $p{=}\tfrac12$ holds as the best point:
\begin{center}
    \small
    \vspace{-1.3em}
    \resizebox{\textwidth}{!}{
    \begin{tabular}{lccccc}
        \toprule
        $p$ & $0$ (TSV-M) & $0.25$ & $0.5$ (derived FTSV) & $1.0$ (PAVE-style) & $2.0$ \\
        \midrule
        \model{ViT-B/32} $T{=}8$ & 83.41 (92.05) & 86.47 (95.52) & \textbf{86.81 (96.02)} & 85.92 (95.07) & 84.93 (93.94) \\
        \model{RoBERTa} & 66.35 (78.01) & 71.28 (85.29) & \textbf{73.13 (87.86)} & 68.40 (82.52) & 56.60 (68.16) \\
        \model{Llama-3.2-3B} & 39.37 & 39.67 & \textbf{39.76} & 36.46 & 13.85 \\
        \bottomrule
    \end{tabular}
    }
\end{center}
The improvement of $p{=}\tfrac12$ over $p{=}1$ is $0.95$ normalized points on vision, $5.34$ on GLUE, and $3.30$ on MergeBench. The optimum is interior in every setting not at either endpoint. On the generative benchmark, however, the margin over $p{=}\tfrac14$ is only $0.09$ points, indicating an agreement with the theoretical prediction rather than as a strong separation from neighboring powers.
Higher powers do not simply strengthen data geometry: the pooled orthogonalization applies $\polar(M\,\Cb^{\,p})\,\Cb^{-p}$ and hence they increasingly relies on inverse powers of the conditioned metric as $p$ grows, with amplification $\kappa(\Cb)^{\,p}$ amplifying the effects of estimation and conditioning.
Experimentally, selecting $p$ on a held-out validation set rather than by derivation, $\tfrac12$ emerges as optimal on all three modalities. On MergeBench, the validation likelihood is minimized at $p{=}\tfrac12$ ($1.640$, against $1.641$ at $p{=}\tfrac14$ and $1.666$ at $p{=}1$). Thus, the value prescribed by the least-squares geometry is also the value selected empirically, providing a direct agreement between derivation and validation. 
\paragraph{Generative merging (Q5).}
\label{sec:p_derivation}
\renewcommand{\arraystretch}{1.4}
\begin{table}[t]
    \centering
    \vspace{-1.2em}
    \caption{Scores per-pillar, every green row at the derived exponent $p{=}\tfrac12$ and every row at its own validation-selected $\alpha$. The input metric's gain is concentrated in math, and the output metric's in safety, at the cost of coding for both \method{F-TSV}.}
        \begin{tabular}{cl cccccc}
            \toprule
            & \multirow{2}{*}{\textbf{Method}} & \multicolumn{6}{c}{\model{Llama-3.2-3B} --- MergeBench} \\
            \cmidrule(lr){3-8}
            & & Instruct. & Math & Coding & Multiling. & Safety & \textbf{Avg.} \\
            \cmidrule(r){2-2} \cmidrule(lr){3-3}\cmidrule(lr){4-4}\cmidrule(lr){5-5}\cmidrule(lr){6-6}\cmidrule(lr){7-7}\cmidrule(lr){8-8} \rowcolor{gray!15}
            \cellcolor{white} & \method{Pre-trained} & $12.68$ & $29.95$ & $27.03$ & $40.13$ & $29.18$ & $27.79$ \\
            \rowcolor{gray!15} \cellcolor{white} \multirow{-2}{*}{\small \vertical{Base}} & \method{Finetuned} & $39.41$ & $57.47$ & $44.23$ & $40.92$ & $76.30$ & $51.66$ \\
            \cdashlinelr{1-8}
            \multirow{3}{*}{\small \vertical{Fixed}} & \method{Model Soup} & $14.75$ & $42.38$ & $37.95$ & $41.81$ & $36.60$ & $34.70$ \\
            & \method{Task Arithmetic} & $13.30$ & $41.24$ & $37.89$ & $41.76$ & $37.78$ & $34.39$ \\
            & \method{TSV-M} & $20.20$ & $50.42$ & $\mathbf{40.41}$ & $\mathbf{41.87}$ & $44.34$ & $39.45$ \\
            \cdashlinelr{1-8}
            & \method{ESM} & $18.30$ & $53.30$ & $39.56$ & $41.19$ & $44.92$ & $39.45$ \\
            \rowcolor{mygreen!50} \cellcolor{white} & \method{\textbf{Func. TSV}} & $\mathbf{21.14}$ & $\mathbf{57.01}$ & $36.46$ & $40.30$ & $44.10$ & $\mathbf{39.80}$ \\
            \rowcolor{mygreen!50} \cellcolor{white} \multirow{-3}{*}{\small \vertical{Data}} & \method{\textbf{Func. TSV}}$^{2}$ & $20.53$ & $56.33$ & $34.53$ & $40.41$ & $\mathbf{46.84}$ & $39.73$ \\
            \bottomrule
        \end{tabular}
    \label{tab:mergebench}
    \vspace{-1.0em}
\end{table}
Here, we ask whether the same previous constructions carry over to generative merging, where the inputs and downstream computations differ substantially. Table~\ref{tab:mergebench} reports five \model{Llama-3.2-3B} MergeBench experts, scored on all five pillars, see Appendix~\ref{app:gemma} for results on \model{Gemma-2-2B}.
At the derived $p = \frac12$, \ftsv{} reaches $39.80$ and \ftsvtwo{} $39.73$, compared with $39.45$ for both \method{TSV-M}, \method{ESM}. The margin, $+0.35$, is an order of magnitude smaller than on vision or GLUE, so we read it as evidence that the data-induced geometry remains applicable to generative merging rather than as a competitive result.
The per-pillar results explain where this transfer comes from: relative to \method{TSV-M}, the input metric gains $+6.60$ on math and $+0.94$ on instruction, while giving back $3.95$ on coding, $1.57$ on multilingual, and $0.24$  on safety. Adding the output metric shifts the balance rather than improving it: \ftsvtwo{} recovers $0.68$ of the math gain and $0.61$ of the instruction gain, adds $+2.74$ in safety, and incurs a further $1.93$ loss in coding. The common pattern is therefore not a uniform improvement across all capabilities, but a redistribution of performance along the geometry captured by the metric. The gains concentrate on the likelihood- and reasoning-oriented pillars, while coding, evaluated through sampled program synthesis, moves in the opposite direction.
\section{Conclusion}
\label{sec:conclusion}
We recast spectral model merging around two geometric decisions: \emph{which directions of an update to keep}, and \emph{which count as interfering}, as questions about data rather than parameters, and answered both from the change a merge induces on real activations. Functional TSV alters only \emph{which} subspace survives and \emph{what orthogonality means}, preserving the spectral experts' content, and reduces exactly to Euclidean TSV-M when the metric is mathematically switched off. 

The ablation separates the benefit between the two data-aware decisions prescribed by the objective. In vision tasks, when averaged across nine settings, three backbones, and $8$ to $20$ tasks, metric-guided selection gains only $+0.11$ normalized points, compared with $+4.16$ from pooled-metric orthogonalization. Notably, the latter benefits increase with the number of tasks, while the former does not. The main effect of replacing Euclidean geometry is therefore not to identify better directions for each task in isolation, but to prevent the directions retained by different tasks from interfering with the data they actually encounter.
This extends beyond vision, the same control in GLUE and in generative merging places orthogonalization ahead in both cases, indicating that this ordering is consistent across different scenarios.
The geometry carries to generative merging as well, where both variants place above data-free and data-driven baselines, while the smaller margin suggests that its effect should be understood primarily as a geometric correction rather than a decisive gain.

\section{Acknowledgments}
This work is supported through the MUR FIS2 grant n. FIS-2023-00942 “NEXUS” (cup B53C25001030001).
%
%
We thank and acknowledge ISCRA for awarding this project access to the LEONARDO supercomputer, owned by the EuroHPC Joint Undertaking, hosted by CINECA (Italy).
We acknowledge the EuroHPC Joint Undertaking for awarding this project access to the EuroHPC supercomputer LEONARDO, hosted by CINECA (Italy) and the LEONARDO consortium through an EuroHPC Development Access call.
The authors declare no competing interests.

\newpage

\bibliography{iclr2027_conference}
\bibliographystyle{iclr2027_conference}

\appendix

\newpage

\section*{Reproducibility statement}

All merging methods, the three evaluation pipelines, configurations and sweep definitions are released. \S\ref{sec:setup} states the protocol; Appendix~\ref{app:conditioning} states the exact conditioning recipe; Appendix~\ref{app:details} states the backbones and expert checkpoints, splits, statistics budgets, rank budgets, coefficient grids, observability rules and evaluation harnesses, together with the selected coefficient of every reported cell (Table~\ref{tab:alphas}). Statistics are cached keyed by task, model, split and budget, and every ablation cell reuses the same caches, so differences between cells reflect the merging rule alone. Every results file records the resolved merger configuration, the selected coefficient and the code revision that produced it. Proposition~\ref{prop:nesting} provides an executable correctness check: a run with a metric switched off must reproduce the parent method exactly, which we verify in \S\ref{sec:mechanism} and Appendix~\ref{app:nesting}.

\section*{Ethics statement}

This work merges publicly available finetuned checkpoints and evaluates them on public benchmarks, using unlabeled examples from each task's own training split solely to estimate second-order statistics; no new data is collected, and no human subjects are involved. One merged Llama expert is a safety alignment expert, and Table~\ref{tab:mergebench} shows that every merging method we evaluate, including ours, recovers part of that expert's safety score. Merging is therefore not a safety-preserving operation by default, and a merged model should not inherit the safety claims of any parent without re-evaluation. Being a single static checkpoint in the base architecture, the merged model inherits the licenses and biases of the base and of every expert combined into it.

\section*{The use of large language models}
\label{sec:llm}

We used generative AI tools (LLM-based coding agents and chat assistants) throughout this project, and we report that use below against the categories of the ICLR policy. We have reviewed all AI-assisted work.

\textbf{Tasks with required disclosure that we did use AI for.}
\emph{Implementing methods.} The released library was written with substantial assistance from an LLM coding agent: the metric machinery (activation-statistics collection for $C$ and the Jacobian-probe estimation of $G$, the whitened selection step and the pooled orthogonalization), the three evaluation pipelines (CLIP on the vision suites, RoBERTa on GLUE, Llama-3.2-3B on MergeBench), the sweep and caching infrastructure, and the scripts that emit the tables and figures of this paper. The baselines we re-implement rather than cite were ported with the same assistance. \emph{Designing and giving feedback on experiments.} The shape of several ablations, the site-attribution grid of \S\ref{sec:sites}, the power sweep of \S\ref{sec:power}, the sample-stability study of Appendix~\ref{app:stability} and the metric-off correctness control implied by Proposition~\ref{prop:nesting}, was discussed with an LLM before being run, and in places proposed by one. \emph{Refining hypotheses and the conceptual framing, and formulating mathematical claims.} The reading of $p=\tfrac12$ as the power at which the orthogonalization step decorrelates tasks \emph{as the data measures it}, the K-FAC reading of the two-sided metric, and the statement that the Euclidean method is the metric-off special case of ours were sharpened in dialogue with an LLM. \emph{Interpreting results.} LLMs were used to summarize sweep outputs, to discuss possible explanations for results we did not expect, and to argue against those explanations. \emph{Translation.} Parts of the manuscript were first drafted in the authors' main language and rendered into English with LLM assistance.

\textbf{Tasks with required disclosure that we did not use AI for.} No synthetic data was generated: every dataset is a public benchmark split used as distributed, and no dataset was cleaned or reformatted by an LLM beyond the loading code noted above. The proofs in Appendices~\ref{app:metricpolar}--\ref{app:nesting} and the derivation of \S\ref{sec:jls} were written and checked by the authors; where an LLM contributed to the framing, as stated above, it did not supply a step of a proof that we did not independently verify. Support for qualitative and thematic data analysis is not applicable to this work.

\textbf{Tasks with recommended disclosure.} We additionally used generative AI tools to draft and edit parts of this manuscript for readability, to produce the plotting code behind our figures, to suggest experimental parameters (coefficient grids, rank budgets and statistics budgets, all of which are then selected empirically on validation data, never by the model's suggestion), and for references literature search.

\textbf{Verification.} Every number reported in this paper is the output of a run of the released code under a released configuration; no result, table entry or score was produced by, or copied from, an LLM. Code that produces reported numbers was reviewed by the authors, and the method carries an executable check of its own: Proposition~\ref{prop:nesting} requires a run with the metric switched off to reproduce the parent method, which we verify in \S\ref{sec:mechanism} and Appendix~\ref{app:nesting}. All citations were checked against the cited papers, so that no reference or attributed claim rests on an LLM's output. We take responsibility for the final content of this work, including text, claims and artifacts produced with the aid of AI.

\section{Limitations}
\label{sec:limitations}

Given the experts and the cached statistics, the merge is deterministic, so the uncertainty that matters is over \emph{which} $N$ unlabeled examples define $C$. We measure it directly: five independent draws (Appendix~\ref{app:stability}) give a standard deviation of $0.11$ normalized points for \ftsv{} on \model{ViT-B/32} at $T{=}8$ and $0.48$ on GLUE, against which the pooled-metric orthogonalization's increment is $3.76$ and $7.97$ respectively, that is $33$ and $14$ standard deviations. We consider differences of that size safe, smaller differences are not, and we mark them where they occur. Two sources of uncertainty remain: the first is the statistics sample, i.e.\ \emph{which} $N$ unlabeled examples define $C$: Appendix~\ref{app:stability} measures how far the metric moves when they are redrawn, on one backbone and for the input metric only. The second is the MergeBench evaluation, two of whose pillars are scored by sampled generation: the same weights returned five-pillar averages spanning $0.14$ points across three evaluations (Appendix~\ref{app:details}), which is the resolution at which the MergeBench rows should be read.

\textbf{The margin over the strongest baseline is method-dependent.} Against TSV-M the input metric is worth $+9.9$ normalized points on GLUE, but against RegMean, the strongest baseline there, it is $+2.4$ (Table~\ref{tab:glue}). The two methods also swap ranks by modality: RegMean is competitive on GLUE and clearly behind on vision. We therefore claim a consistent improvement rather than a uniform.

\textbf{The output metric underperforms.} Paired across five draws of the statistics sample with the coefficient pinned at its published value, \ftsvtwo{} exceeds \ftsv{} by $0.15 \pm 0.05$ normalized points on \model{ViT-B/32} ($T{=}8$), positive in all five draws, and by $0.13 \pm 0.76$ on GLUE, where the per-draw difference ranges from $-0.93$ to $+0.93$ and changes sign. The vision margin is therefore real and small; the GLUE margin is not distinguishable from the draw. Allowing each draw to re-select its own coefficient, as the reported cells do, adds enough selector noise to bury the vision effect as well ($0.12 \pm 0.13$). We report the output metric as completing the K-FAC picture at a cost in backward passes that grows with model size, and not as a result the paper rests on. We therefore recommend \ftsv{} by default, and \ftsvtwo{} only when maximizing performance outweighs its additional offline cost, as in safety-critical deployments.

\textbf{Coefficient selection on MergeBench.} A generative benchmark offers no held-out accuracy at a per-$\alpha$ cost, so $\alpha$ is chosen there by teacher-forced validation likelihood (Appendix~\ref{app:details}). The three rows of the MergeBench chain in Table~\ref{tab:alphas} select $\alpha=0.8$, $0.8$ and $1.8$, so its increments combine the change of geometry with a change of coefficient. The coefficients differ because equal $\alpha$ does not give equal steps on the data: on vision and GLUE, TSV-M run at the step of \ftsv{} scores as at its own coefficient (Appendix~\ref{app:preserve}), and on MergeBench the validation likelihood of \ftsv{} stays below the best of TSV-M from $\alpha=0.9$ to $2.0$ (Table~\ref{tab:nllcurve}).

\textbf{Assumptions and scope.} Assumption~\ref{as:kfac} treats input statistics and downstream Jacobians as independent, where that assumption does not hold, the two-sided norm misrepresents the curvature; measured on \model{ViT-B/32} and GLUE, the K-FAC metric reproduces the exact cross-task functional geometry of the merge only at $r=0.30$ and $0.43$ (ceiling $0.98$--$0.99$), mainly because inputs and downstream Jacobians are not independent, a measured, contributing cause of the small gain of \ftsvtwo{}, lefting open space to future work that could account this dependence. As with all task-vector methods, we assume a single shared pretrained model.

\textbf{Data requirements.} \ftsv{} needs only a few unlabeled inputs per task, passed once through the pretrained model (about $10$\,s for GLUE, $21$\,s to $2.2$ minutes on vision), and the merge reads only their second moments $C_t$, which a task owner can release instead of the data. Labels serve only to select $\alpha$, as for every baseline; on vision a single $\alpha=1.6$ stays within $0.5$ validation points of each setting's optimum, whereas GLUE is more sensitive to $\alpha$.

\section{Theory}
\label{sec:theory}
\begin{table}[t]
    \centering
    \caption{\textbf{Notation}. Merging acts per weight matrix, layer indices are removed to simplify notation. The input metric is $C$ and the output metric is $G$ at every stage, raw or conditioned, and a metric enters the merge only through a power $p$ of itself. Decorations are used consistently: $\bar{\cdot}$ pools over tasks, $\hat{\cdot}$ stacks per-task factors,
    $\tilde{\cdot}$ lives in whitened coordinates, $\cdot^{(k)}$ is a task's rank-$k$ block.}
    \label{tab:notation}
    \footnotesize
    \resizebox{1\textwidth}{!}{
        \begin{tabular}{@{}ll@{\hskip 2em}ll@{}}
            \toprule
            $\theta_0,\ \theta_t$ & base weights; expert $t$ & $C_t\in\R^{n\times n}$ & conditioned $\E_{x\sim\mathcal D_t}[xx^{\top}]$ \\
            $\tau_t=\theta_t-\theta_0$ & task vector of expert $t$ & $G_t\in\R^{m\times m}$ & conditioned $\E[J^{\top}J]$ \\
            $\Delta_t\in\R^{m\times n}$ & one weight matrix of $\tau_t$ & $\Cb,\Gb$ & pooled metrics, $\Cb=\frac1T\sum_t\unitop(C_t)$ \\
            $\Delta_t^{(k)}=U_t^{(k)}S_t^{(k)}V_t^{(k)\top}$ & task $t$'s rank-$k$ block & $C_t^{\,p},\ G_t^{\,p_g}$ & whitening factors (powers of a metric) \\
            $\Delta,\ \Delta^{\star}$ & merged update; $\star$ marks the solution & $p,\ p_g$ & whitening powers, derived $p=p_g=\tfrac12$ \\
            $\theta=\theta_0+\alpha\Delta^{\star}$ & the merged model & $r,\ r_g$ & conditioning ranks, absolute or a fraction of the width \\
            $\hat U,\hat s,\hat V$ & per-task factors, stacked & $\unitop(\cdot)$ & rescaling to unit mean eigenvalue \\
            $L,\ R$ & the stacks, orthogonalized & $\cond(\cdot\,;r)$ & the conditioning map of Appendix~\ref{app:conditioning} \\
            $T,\ k$ & tasks; rank kept per task and layer & $\kappa(\cdot)$ & condition number of a conditioned metric \\
            $N$ & unlabeled examples per task & $n_p$ & Gaussian VJP probes per batch \\
            $\alpha$ & merging coefficient & $\polar(\cdot)$ & nearest orthonormal factor \\
            \bottomrule
        \end{tabular}
    }
\end{table}
We first state the assumptions underlying the derivation of the metric, which is well-defined for any PSD $C_t$ and $G_t$:
\begin{assumption}[Local quadratic losses]\label{as:quad}
    $\mathcal L_t(\theta)\approx \mathcal L_t(\theta_t)+\tfrac12(\theta-\theta_t)^{\top}H_t(\theta-\theta_t)$ about each expert, with $H_t\succeq0$.
\end{assumption}
\begin{assumption}[K-FAC factorization]\label{as:kfac}
    Per layer, $\E[(xx^{\top})\otimes(J^{\top}J)]\approx C_t\otimes G_t$ is exact whenever $J$ is constant over the data \citep{martens2015kfac}.
\end{assumption}
\begin{assumption}[Spectral sparsity]\label{as:rank}
$   \Delta=\sum_t\Delta_t^{(k)}$ with $\rank(\Delta_t^{(k)})\le k=\lfloor\min(m,n)/T\rfloor$.
\end{assumption}
\begin{assumption}[Isotropic representation curvature]\label{as:iso}
The Hessian of task $t$'s loss with respect to the final representation is isotropic, $H_\ell=\nabla_f^2\ell_t\propto I$: task $t$'s loss is measured by the squared change of the final representation.
\end{assumption}
Under Assumptions~\ref{as:quad}, \ref{as:kfac} and \ref{as:iso} the per-layer term of \eqref{eq:jls} is the two-sided functional norm (Appendix~\ref{app:kron}). Assumption~\ref{as:iso} makes the sensitivity metric task-loss-agnostic and label-free, and Gaussian vector-Jacobian products estimate it efficiently (\S\ref{sec:ftsv2}). The unit-mean rescaling of Appendix~\ref{app:conditioning} removes a positive scalar in $H_\ell$, but not its anisotropy, which we leave to future work (see Appendix \ref{sec:limitations}).

\begin{proposition}[Nearest metric-orthonormal factor]
\label{prop:polar}
Let $A\succ0$ be a conditioned metric and $p>0$. Then $R^{\star}=\polar(\hat V^{\top}A^{\,p})A^{\,-p}$ minimizes $\norm{(R-\hat V^{\top})A^{\,p}}_F$ subject to $R\,A^{\,2p}R^{\top}=I$, and attains it. Two corners: at $p=\tfrac12$ the constraint is $A$-orthonormality, the non-interference criterion of \eqref{eq:interference}; at $A=I$ the solution is the Euclidean polar factor. Taking $A=\Cb$ gives \eqref{eq:metricortho}, and the mirror statement for columns orthonormal in $\Gb$ gives \eqref{eq:leftortho}.
\end{proposition}

\begin{proposition}[The orthogonalization fixes the power]
    \label{prop:exponent}
    Let $\Cb\succ0$ be the conditioned pooled metric with $\Cb\neq I$, and write $\norm{M}_{\Cb}^2:=\tr(M\Cb M^{\top})$ for the quadratic form of \eqref{eq:interference}, which equals $\E_x\norm{Mx}^2$ when $\Cb$ is the second moment of $x$. Then (i)~the constraint $R\,\Cb^{\,2p}R^{\top}=I$ of \eqref{eq:metricortho} defines the same set of matrices as $\Cb$-orthonormality, $R\,\Cb\,R^{\top}=I$, if and only if $p=\tfrac12$; and (ii)~the objective $\norm{(R-\hat V^{\top})\Cb^{\,p}}_F^2$ equals $\norm{R-\hat V^{\top}}_{\Cb}^2$ for every $R$ if and only if $p=\tfrac12$. The mirror statements hold on the output side with $\Gb$ and $p_g$.
\end{proposition}

Selection, by contrast, uses $C_t^{\,p}$ only to rank directions and never rescales the kept block \eqref{eq:select}; it is well defined at every $p>0$ and constrains none. The power is therefore fixed by the orthogonalization alone, the step that Figure~\ref{fig:mechanism} finds to carry the gain, and \S\ref{sec:power} measures it there.

\begin{lemma}[Range alignment]
    \label{lem:range}
    With $G_t^{\,p_g}\Delta_t C_t^{\,p} = \tilde U \tilde S \tilde V^{\top}$ and $Q_t = \mathrm{QR}(C_t^{\,p} \tilde V_k)$, $\ \mathrm{range}(\Delta_t Q_t)=\mathrm{range}(G_t^{\,-p_g}\tilde U_k)$: selecting the top-$k$ \emph{input} directions of the doubly whitened update confines the block's outputs to the top-$k$ whitened \emph{output} subspace, so no separate left truncation is needed (Appendix~\ref{app:range}).
\end{lemma}

\begin{proposition}[Nested exactness]
    \label{prop:nesting}
    Per layer: (i)~if $r_g=0$, $p_g=0$, or no output statistic was observed, \ftsvtwo{} coincides with \ftsv{}; (ii)~if in addition $r=0$, $p=0$, or no input statistic was observed, it coincides with TSV-M, i.e.\ with \eqref{eq:tsvm}. In every case the switched-off metric enters as the exact identity, $A^{\,p}=A^{\,-p}=I$, so the reductions are identities of the computation up to floating-point roundoff, not limits: the Euclidean method is a corner of the family along either axis, rank or power.
\end{proposition}

Proofs are in Appendices~\ref{app:metricpolar}--\ref{app:nesting}; Appendix~\ref{app:nesting} also verifies the reductions numerically. Proposition~\ref{prop:nesting} makes every ablation cell of \S\ref{sec:mechanism} an interpolation between two published methods: a site that keeps no metric executes, at that layer, exactly the TSV-M method.

\begin{algorithm}[h]
\caption{Functional TSV (\ftsv{}) and its two-sided extension (\ftsvtwo{}).
Dropping every $G_t$ gives \ftsv{}; dropping every $C_t$ as well gives TSV-M
(Proposition~\ref{prop:nesting}).}
\label{alg:ftsv}
\begin{algorithmic}[1]
\REQUIRE base $\theta_0$; experts $\theta_1..\theta_T$; $N$ unlabeled
examples per task; conditioning rank $r$ and power $p$ on the input side and,
two-sided, $(r_g,p_g,n_p)$ on the output side; ranks are absolute or a
fraction of the layer width
\STATE \textbf{Statistics.} For each task $t$: accumulate $C_t$ over hooked
layer inputs (token rows, padding masked); two-sided, accumulate $G_t$ from
$n_p$ Gaussian VJP probes per batch; on a causal LM, the two factors of the
token-embedding matrix (Appendix~\ref{app:embed}). Cache to disk.
\FORALL{weight matrices $W\in\R^{m\times n}$}
  \STATE $k\gets\max(1,\lfloor\min(m,n)/T\rfloor)$
  \STATE $\Cb\gets\cond(\frac1T\sum_t\unitop(C_t);r)$,\;
    $\Gb\gets\cond(\frac1T\sum_t\unitop(G_t);r_g)$
    \hfill$\triangleright$ pooled from the raw statistics
  \FORALL{tasks $t=1..T$}
    \STATE $C_t\gets\cond(C_t;r)$ if $C_t$ observed, else $I$;\quad
      likewise $G_t\gets\cond(G_t;r_g)$
    \IF{either metric is active}
      \STATE $(\cdot,\cdot,\tilde V)\gets\mathrm{SVD}(G_t^{\,p_g}\Delta_tC_t^{\,p})$;\;
        $Q_t\gets\mathrm{QR}(C_t^{\,p}\tilde V_{[:,1:k]})$
        \hfill$\triangleright$ the metric selects the subspace
      \STATE $(U_t^{(k)},S_t^{(k)},V_t^{(k)})\gets$ SVD of $\Delta_tQ_tQ_t^{\top}$
        \hfill$\triangleright$ content undistorted, \eqref{eq:select}
    \ELSE
      \STATE $(U_t^{(k)},S_t^{(k)},V_t^{(k)})\gets$ top-$k$ SVD of $\Delta_t$
        \hfill$\triangleright$ equal to TSV-M
    \ENDIF
  \ENDFOR
  \STATE stack the blocks into $\hat U,\hat s,\hat V^{\top}$
  \STATE $R\gets\polar(\hat V^{\top}\Cb^{\,p})\,\Cb^{\,-p}$ if $C$ active else
    $\polar(\hat V^{\top})$ \hfill$\triangleright$ \eqref{eq:metricortho}
  \STATE $L\gets\Gb^{\,-p_g}\polar(\Gb^{\,p_g}\hat U)$ if $G$ active else
    $\polar(\hat U)$ \hfill$\triangleright$ \eqref{eq:leftortho}
  \STATE merged layer $\gets L\diag(\hat s)R$
\ENDFOR
\STATE non-matrix parameters $\gets$ mean of task vectors
\RETURN $\theta_0+\alpha\Delta^{\star}$, $\alpha$ selected on validation data
\end{algorithmic}
\end{algorithm}

\section{JLS approximation ladder}
\label{app:algorithm}

Algorithm~\ref{alg:ftsv} states the two methods. Table~\ref{tab:ladder_theory} places them among the instances of \eqref{eq:jls}. Each row fixes the two ingredients of the objective: the curvature $H_t$ that prices a deviation from expert $t$, and the constraint $\Omega$ on the merged update. The first four rows impose no constraint and solve \eqref{eq:jls} exactly, in closed form or, for MaTS \citep{tam2024mats}, by conjugate gradients. They differ only in how much structure $H_t$ keeps, from the identity through the diagonal Fisher to Kronecker factors of a layer's input statistics (RegMean \citet{jin2023regmean}) and of both of its sides (MaTS \citet{tam2024mats}). With the coefficient $\alpha$ of \S\ref{sec:notation}, the first row is task arithmetic, and $\alpha=1$ gives weight averaging. The exact solutions of the Kronecker rows multiply by the inverse of a sum of estimated statistics, $(\sum_tC_t)^{-1}$ for RegMean, and keep no per-task low-rank structure.
The last three rows replace $\Omega=0$ by the per-task spectral constraint of Assumption~\ref{as:rank}, and each keeps the curvature of an unconstrained row: TSV-M \citep{gargiulo2025tsv} the identity, \ftsv{} the $C_t\otimes I$ of RegMean, and \ftsvtwo{} the two-sided Kronecker form of MaTS, with the label-free sensitivity $G_t=\E[J^{\top}J]$ of \S\ref{sec:ftsv2} as its output factor. The rank constraint only sets the spectral budget: the polar rule follows from non-interference and the nearest-factor criterion, and exactly solves the factor-level problem $\min_R\norm{(R-\hat V^{\top})\Cb^{\,p}}_F^2$ subject to $R\,\Cb^{\,2p}R^{\top}=I$, whose exact solution is $R^{\star}=\polar(\hat V^{\top}\Cb^{\,p})\,\Cb^{\,-p}$ (Proposition~\ref{prop:polar}), but not \eqref{eq:jls}, which motivates the geometry.
\begin{table}
\centering
\caption{Merging methods as instances of \eqref{eq:jls}. The spectral rows use merge rules motivated by the geometry of Proposition~\ref{prop:polar}. Kronecker factors are stated for column-stacked $\vecop(\Delta)$, where $\vecop(\Delta)^{\top}(C\otimes G)\vecop(\Delta)=\norm{G^{1/2}\Delta C^{1/2}}_F^2$ (as in Appendix~\ref{app:kron}).}
\label{tab:ladder_theory}
\small
\begin{tabular}{@{}llll@{}}
\toprule
$H_t$ & $\Omega$ & Merge rule & Method \\
\midrule
$I$ & $0$ & $\Delta=\frac1T\sum_t\tau_t$ & Task Arithmetic \\
$\diag(F_t)$ & $0$ & per-coordinate weighted mean & Fisher merging \\
$C_t\otimes I$ (per layer) & $0$ &
  $\Delta=(\sum_t\Delta_tC_t)(\sum_tC_t)^{-1}$ & RegMean \\
block K-FAC & $0$ & CG solve of the normal equations & MaTS \\
$I$ & spectral, per task & select $k$ dirs/task $+$ orthogonalize & TSV-M \\
$C_t\otimes I$ & spectral, per task & \S\ref{sec:ftsv} & \ftsv{} (ours) \\
$C_t\otimes G_t$ & spectral, per task & \S\ref{sec:ftsv2} & \ftsvtwo{} (ours) \\
\bottomrule
\end{tabular}
\end{table}
\section{Functional norm and its Kronecker form}
\label{app:kron}

\textbf{One-sided identity.} For $\Delta\in\R^{m\times n}$ and $x\sim\mathcal D$ with $C=\E[xx^{\top}]$,
\[
\E\norm{\Delta x}^2=\E\big[\tr(\Delta^{\top}\Delta\,xx^{\top})\big]
=\tr(\Delta^{\top}\Delta C)=\tr(C^{1/2}\Delta^{\top}\Delta C^{1/2})
=\norm{\Delta C^{1/2}}_F^2 .
\]
The functional size of an update is thus its Frobenius norm after right-whitening: input directions with large data energy are stretched, directions the data never visits are shrunk.

\textbf{Two-sided approximation.} With $J(x)=\partial f/\partial y$ the Jacobian of the final representation with respect to the layer output, a perturbation
$\delta y=\Delta x$ changes $f$ by $J(x)\Delta x$ to first order, so
\[
\E\norm{J(x)\Delta x}^2
=\E\big[\tr(\Delta^{\top}J^{\top}J\,\Delta\,xx^{\top})\big]
\;\approx\;\tr(\Delta^{\top}G\Delta C)=\norm{G^{1/2}\Delta C^{1/2}}_F^2 ,
\]
where the approximation is Assumption~\ref{as:kfac}, exact whenever $J$ is constant over the data.

\textbf{Kronecker form.} For column-stacked vectorization, $(A\otimes B)\vecop(X)=\vecop(BXA^{\top})$, so for symmetric $C,G$,
\[
\vecop(\Delta)^{\top}(C\otimes G)\vecop(\Delta)=\vecop(\Delta)^{\top}
\vecop(G\Delta C)=\tr(\Delta^{\top}G\Delta C)=\norm{G^{1/2}\Delta C^{1/2}}_F^2 .
\]
For $y=Wx$ we have $y=(x^{\top}\otimes I_m)\vecop(W)$, hence the Gauss--Newton block restricted to $\vecop(W)$ is $\E[(xx^{\top})\otimes(J^{\top}J)]\approx C\otimes G$, whose quadratic form is the two-sided norm above.

\section{Metric conditioning}
\label{app:conditioning}

\textbf{Estimation.} For every observable layer, the $N$ unlabeled examples of task $t$ are passed once through the pretrained model $\theta_0$, and the second moment of the layer's input is accumulated over the $M$ token rows they produce, with padding masked: $C_t=\frac1M\sum_{i=1}^{M}x_ix_i^{\top}$. On vision and GLUE $N=512$; Table~\ref{tab:budgets} lists every budget. Because the estimate comes from a small sample, every metric is conditioned before the merge uses it.

\textbf{The conditioning map.} Rescale the statistic to unit mean eigenvalue and write its eigendecomposition as $\sum_j\lambda_jw_jw_j^{\top}$, with $\frac1n\sum_j\lambda_j=1$. We keep the top $r$ eigendirections and floor the remaining tail to the isotropic level,
\begin{equation}
  C_t \;\propto\; I \;+\; \sum_{j \le r} (\lambda_j - 1)\, w_j w_j^{\top},
  \label{eq:conditioning}
\end{equation}
and rescale the result to unit mean eigenvalue. Conditioning moves eigenvalues only, and we keep calling the result $C_t$ throughout the paper. The same map, $\cond(\cdot\,;r)$ in Algorithm~\ref{alg:ftsv}, conditions every metric the method uses, per-task or pooled, input or output: $C_t$, $\Cb$, $G_t$ and $\Gb$. Its only hyperparameter is the rank $r$, absolute on the vision and GLUE encoders and a fraction of each layer's width on \model{Llama-3.2-3B} (Table~\ref{tab:budgets}). Given a raw statistic $A\in\R^{n\times n}$, a rank $r$ and a power $p>0$, the map
\begin{enumerate}\itemsep0pt
    \item symmetrizes, $A\gets\frac12(A+A^{\top})$, and returns identity factors if $\tr A\le0$, a degenerate statistic;
    \item rescales to unit mean eigenvalue, $A\gets A\cdot n/\tr(A)$;
    \item eigendecomposes $A=U\Lambda U^{\top}$ once, keeps the top $r$ eigenvalues, sets the other $n-r$ to $1$, and rescales to unit mean eigenvalue;
    \item returns the whitening factor $A^{\,p}=U\diag(\lambda^{p})U^{\top}$ and, where needed, its inverse $A^{\,-p}=U\diag(\lambda^{-p})U^{\top}$.
\end{enumerate}
Every power follows from the single eigendecomposition of step~3, and for $r\ll n$ each factor is a rank-$r$ update of a scaled identity.

\textbf{Why the tail is floored.} The merge uses a metric only through its powers. Truncating the tail to zero would leave $A^{\,-p}$ undefined; flooring it at the isotropic level keeps the tail's mass and every eigenvalue away from zero. This bounds the only inverse factors the merge forms, $\Cb^{\,-p}$ in \eqref{eq:metricortho} and, for \ftsvtwo{}, $\Gb^{\,-p_g}$ in \eqref{eq:leftortho}: with unit mean eigenvalue, each stretches a direction by at most $\kappa^{p}$, with $\kappa$ the condition number of the conditioned metric. The per-task metrics $C_t$ and $G_t$ enter only through positive powers, in the selection step, and are never inverted. At $r=0$ every eigenvalue is $1$ and $A^{\,p}=A^{\,-p}=I$ exactly: this is the TSV-M anchor of Proposition~\ref{prop:nesting}, reached along the same axis $r$ that controls the conditioning. The power axis reaches the same anchor at $p=0$, where the factors are also returned as the exact identity.

\textbf{Measured condition numbers.} On \model{ViT-B/32} ($T{=}8$), over the $48$ observable matrices, the conditioned $\Cb$ has a median $\kappa$ of $260$ (range $90$ to $1456$), so the amplification at $p=\tfrac12$ is $16$ in the median and at most $38$. When the statistics sample is redrawn, $\kappa$ moves by $0.35\%$ in the median and by at most $4.1\%$ (Appendix~\ref{app:stability}).

\textbf{The rank.} Sweeping $r\in\{0,32,64,128,256,512\}$ with everything else fixed, the normalized test accuracy is $92.04$, $95.55$, $95.63$, $96.02$, $95.70$ and $95.93$ on \model{ViT-B/32} ($T{=}8$), and $78.02$, $85.95$, $85.61$, $87.86$, $85.47$ and $85.87$ on GLUE. At $r=32$ the metric already delivers $88\%$ (vision) and $80\%$ (GLUE) of the gain over $r=0$ that $r=128$ delivers. The test score peaks at $r=128$ on both benchmarks and is not monotone above it. On the validation split, which selects each cell's $\alpha$, the five ranks from $32$ to $512$ lie within $0.15$ normalized points of one another on \model{ViT-B/32}, highest at $r=32$; on GLUE, validation also peaks at $r=128$. At $r=0$ the metric is the identity, and the sweep reproduces TSV-M.

\section{Proof of Proposition~\ref{prop:polar}}
\label{app:metricpolar}

\textbf{Euclidean case.} $\polar(Z)=\tilde U\tilde V^{\top}$ from
$Z=\tilde U\tilde S\tilde V^{\top}$ minimizes $\norm{Z-X}_F$ over $X$ with
orthonormal rows (equivalently columns).

\textbf{Metric case.} We seek $R$ nearest to $\hat V^{\top}$ in the functional
norm subject to $R\,\Cb^{\,2p}R^{\top}=I$:
$\min_{R}\norm{(R-\hat V^{\top})\Cb^{\,p}}_F$. Substituting
$\tilde R=R\,\Cb^{\,p}$ turns the constraint into $\tilde R\tilde R^{\top}=I$
and the objective into $\norm{\tilde R-\hat V^{\top}\Cb^{\,p}}_F$, whose
minimizer is $\tilde R=\polar(\hat V^{\top}\Cb^{\,p})$ by the Euclidean case.
Mapping back gives $R=\polar(\hat V^{\top}\Cb^{\,p})\Cb^{\,-p}$ and
$R\,\Cb^{\,2p}R^{\top}=\tilde R\tilde R^{\top}=I$, which is
\eqref{eq:metricortho}; at $p=\tfrac12$ the constraint reads $R\,\Cb\,R^{\top}=I$.
The output-side rule \eqref{eq:leftortho} follows from the mirror substitution
$\tilde L=\Gb^{\,p_g}L$. At $\Cb=I$ (equivalently $p\to0$) both reduce to the
Euclidean polar factor. \hfill$\square$

\begin{remark}
A row of $R$ has unit \emph{data} norm, so along directions the data barely
excites its Euclidean length may exceed one, intentionally, since weight mass
there is functionally cheap. The isotropic tail caps the effect at
$\kappa(\Cb)^{p}$ (Appendix~\ref{app:conditioning}), and \S\ref{sec:power} is
where that trade stops being free.
\end{remark}

\section{Proof of Proposition~\ref{prop:exponent}}
\label{app:exponent}

Both statements reduce to one fact: two symmetric positive definite matrices
with the same quadratic form are equal. Write $A=\Cb^{\,2p}$ and $B=\Cb$.

\textbf{Statement (ii).} $\norm{M\Cb^{\,p}}_F^2=\tr(MAM^{\top})$ and
$\norm{M}_{\Cb}^2=\tr(MBM^{\top})$. If the two agree for every $M$, taking
$M=w^{\top}$ for an arbitrary vector $w$ gives $w^{\top}Aw=w^{\top}Bw$ for all
$w$, hence $A=B$; the converse is immediate.

\textbf{Statement (i).} Suppose every $R\in\R^{q\times n}$ ($q\le n$) with
$RAR^{\top}=I$ also satisfies $RBR^{\top}=I$. Take any $w$ with $w^{\top}Aw=1$,
extend it to $q$ vectors that are $A$-orthonormal, possible since $q\le n$, and
stack them as the rows of $R$. Then $RAR^{\top}=I$, so $RBR^{\top}=I$, so
$w^{\top}Bw=1$: every $A$-unit vector is a $B$-unit vector, and by scaling
$w^{\top}Aw=w^{\top}Bw$ for all $w$, hence $A=B$. The converse is immediate.

\textbf{$A=B$ forces $p=\tfrac12$.} $\Cb$ is symmetric with eigenvalues
$\lambda_j>0$, and $\Cb^{\,2p}$ has the same eigenvectors with eigenvalues
$\lambda_j^{2p}$, so $\Cb^{\,2p}=\Cb$ iff $\lambda_j^{2p}=\lambda_j$ for every
$j$. Since $\Cb\neq I$ some $\lambda_j\neq1$, and for that $j$ the equation
holds iff $2p=1$.

\textbf{Output side.} Substitute $\tilde L=\Gb^{\,p_g}L$ as in
Appendix~\ref{app:metricpolar} and use $\norm{u}_{\Gb}^2=u^{\top}\Gb u$, the
quadratic form of $\langle u,u'\rangle_{\Gb}$ in \S\ref{sec:ftsv2}; the two
statements become their mirror images and the same argument gives
$p_g=\tfrac12$. \hfill$\square$

\begin{remark}
When $\Cb=I$ (conditioning rank $r=0$, Appendix~\ref{app:conditioning}) every
$p$ gives the same Euclidean polar factor and nothing is fixed. The power is
determined exactly when there is a data geometry to determine it against,
which is the Euclidean anchor of Proposition~\ref{prop:nesting} seen from the
other side.
\end{remark}

\section{Proof of Lemma~\ref{lem:range}}
\label{app:range}

Let $G_t^{\,p_g}\Delta_tC_t^{\,p}=\tilde U\tilde S\tilde V^{\top}$ and
$Q_t=\mathrm{QR}(C_t^{\,p}\tilde V_k)$, so
$\mathrm{range}(Q_t)=\mathrm{range}(C_t^{\,p}\tilde V_k)$. Then
\[
\Delta_t\,C_t^{\,p}\tilde V_k
=G_t^{\,-p_g}\big(G_t^{\,p_g}\Delta_tC_t^{\,p}\big)\tilde V_k
=G_t^{\,-p_g}\tilde U\tilde S\tilde V^{\top}\tilde V_k
=G_t^{\,-p_g}\tilde U_k\tilde S_k ,
\]
hence $\mathrm{range}(\Delta_tQ_t)=\mathrm{range}(G_t^{\,-p_g}\tilde U_k)$: keeping
the top-$k$ \emph{input} directions of the doubly whitened update confines the
block's outputs to the top-$k$ whitened output subspace mapped back to raw
coordinates. Selecting on one side selects both. \hfill$\square$

\section{Proof of Proposition~\ref{prop:nesting}}
\label{app:nesting}

At $r=0$ the conditioned metric is the identity exactly
(Appendix~\ref{app:conditioning}): no eigendirection is retained, every
eigenvalue sits at the isotropic level, and $C_t^{\,p}=C_t^{\,-p}=I$ for every
$p$; at $p=0$ the factors are the identity for every conditioned metric; and a
layer whose statistic was never observed is assigned the identity. With the
output side off, every $G$ factor in Algorithm~\ref{alg:ftsv} is a
multiplication by $I$, so each layer executes the one-sided computation. With
both sides off, the whitened update is $\Delta_t$ itself, $Q_t$ spans its
top-$k$ right singular subspace, $\Delta_tQ_tQ_t^{\top}$ is its rank-$k$
truncation, and both polar steps are Euclidean, which is \eqref{eq:tsvm}. The
two computations differ only by the extra factorizations of the metric path (a
QR of an already orthonormal basis and a second SVD of the same block), which
agree in exact arithmetic. \hfill$\square$

\textbf{Verification.} On the reported cells, running \ftsvtwo{}
with the output metric kept at no site returns $86.81$ ($96.02$) on
\model{ViT-B/32} ($T{=}8$) and $73.13$ ($87.86$) on GLUE, the \ftsv{} rows of
Tables~\ref{tab:vision} and \ref{tab:glue} in every stored digit (the
\method{none} rows of the output panels of Tables~\ref{tab:sites} and
\ref{tab:sites_glue}). On the aggregation itself, in double precision on
synthetic low-rank task vectors (five tasks, three layers of widths $64$ to
$128$), the three routes to the identity input metric, $p=0$, $r=0$, and no
observed statistic, agree with the plain TSV-M aggregation to a relative
difference below $2\times10^{-14}$. On MergeBench, where the evaluation is not
deterministic (Appendix~\ref{app:details}), the $p{=}0$ cell of
Table~\ref{tab:power_full} ($39.37$) and the TSV-M row of Table~\ref{tab:mergebench} ($39.45$) are two evaluations of the same weights, up to floating-point rounding.

\section{\ftsv{} preserves the expert update}
\label{app:preserve}

The abstract states that \ftsv{} uses the data metric to choose each task's subspace and to define cross-task orthogonality, while preserving the expert update within the chosen subspace. This appendix states what is preserved, by which step, and what the orthogonalization step changes.

\textbf{Step 1 keeps the expert's own action.} Selection uses the metric only to choose the orthonormal basis $Q_t$ of \eqref{eq:select}, and the kept block is $\Delta_t^{(k)}=\Delta_tQ_tQ_t^{\top}$. Hence
\[
\Delta_t^{(k)}x=\Delta_tx\ \ \text{for every } x\in\mathrm{range}(Q_t),
\qquad
\Delta_t^{(k)}x=0\ \ \text{for every } x\perp\mathrm{range}(Q_t):
\]
on the input directions it keeps, the block responds exactly as the expert does. No power of $C_t$ multiplies the kept content, and the singular values $\hat s_t$ of the block are those of $\Delta_tQ_t$, the expert restricted to the chosen subspace, and it holds for every power $p$ and every conditioning rank $r$.

\textbf{Step 2 keeps the singular values and the subspaces, and renormalizes the directions.} The pooled orthogonalization \eqref{eq:metricortho} acts on the stacked factors $\hat U,\hat s,\hat V$ of the $T$ blocks. Since $\polar(Z)=(ZZ^{\top})^{-1/2}Z$ whenever $Z$ has full row rank, as $Z=\hat V^{\top}\Cb^{\,p}$ does,
\[
R=\polar\big(\hat V^{\top}\Cb^{\,p}\big)\,\Cb^{\,-p}=\big(\hat V^{\top}\Cb^{\,2p}\hat V\big)^{-1/2}\hat V^{\top},
\]
whereas TSV-M uses 
\[
\ \polar(\hat V^{\top})=\big(\hat V^{\top}\hat V\big)^{-1/2}\hat V^{\top}.
\]
Both are the symmetric orthogonalization of the selected directions; \ftsv{} computes their Gram matrix in the data metric instead of the Euclidean one. Three properties of the merged update $\Delta^{\star}=L\diag(\hat s)R$, with $L=\polar(\hat U)$, follow.
\begin{itemize}\itemsep0pt
    \item \emph{It stays in the chosen subspaces.} Every row of $R$ is a combination of the selected directions, so $\Delta^{\star}$ reads its input only from $\mathrm{range}(\hat V)$, the span of the directions selected in Step~1, and writes only into $\mathrm{range}(\hat U)$, as in TSV-M.
    \item \emph{It keeps the singular values $\hat s$.} $\Delta^{\star}\Cb^{\,p}=L\diag(\hat s)\polar(\hat V^{\top}\Cb^{\,p})$ is $\diag(\hat s)$ between two orthonormal factors, so $\hat s$ are exactly the singular values of the merged update in the whitened coordinates of the data metric, where TSV-M has them in weight coordinates. On the merged layers of all nine vision settings and GLUE, $\norm{\Delta^{\star}\Cb^{1/2}}_F/\norm{\hat s}_2=1.000$ (median over layers).
    \item \emph{It changes the Euclidean length of the directions.} Each row $r$ of $R$ has unit length in the data metric, $\norm{r\,\Cb^{\,p}}=1$, so its Euclidean length lies between $\lambda_{\max}(\Cb)^{-p}$ and $\lambda_{\min}(\Cb)^{-p}$. Since $\Cb$ has unit mean eigenvalue, directions that the pooled data excites more than the average are shortened and those it excites less are lengthened, by a factor of at most $\kappa(\Cb)^{p}$, which is $16$ in the median layer of \model{ViT-B/32} and at most $38$ (Appendix~\ref{app:conditioning}). This is what orthonormality in the data metric means (the remark in Appendix~\ref{app:metricpolar}), and it acts even on a single task: at $T=1$, on a layer with $n\le m$, TSV-M returns $\Delta_1$ whereas \ftsv{} returns $\Delta_1\Cb^{\,-p}$.
\end{itemize}

\textbf{The overall change of scale is absorbed by $\alpha$.} Because $\hat s$ is carried in the data metric rather than in weight coordinates, the merged update of \ftsv{} is smaller than that of TSV-M on the data, though not in weight space. Per layer, its size on the pooled inputs, $\tr(\Delta^{\star}C_{\text{mix}}\Delta^{\star\top})^{1/2}$ with $C_{\text{mix}}=\frac1T\sum_tC_t$, is $0.49$ to $0.58$ times that of TSV-M (median over layers, nine vision settings and GLUE), while its Frobenius norm is $1.01$ to $1.26$ times as large. The validated coefficients compensate: $\alpha_{\text{TSV-M}}/\alpha_{\ftsv{}}$ is $0.47$ to $0.60$, within $0.08$ of the size ratio in every setting. This is why the $\alpha$ of \ftsv{} in Table~\ref{tab:alphas} is about twice that of TSV-M, while that of metric-guided selection alone is not. Run at the coefficient that gives it the step of \ftsv{} on the data, TSV-M scores as at its own coefficient: $92.04$, $85.64$ and $78.57$ normalized on \model{ViT-B/32} at $T{=}8$ and $T{=}20$ and on GLUE, against $92.05$, $85.91$ and $78.01$. At their validated coefficients the two methods therefore differ in which directions they keep and in how the kept directions are normalized and combined, not in the size of the step they take on the data.

\textbf{Common-coefficient diagnostic.} Table~\ref{tab:ccd} repeats this comparison in every setting, normalized accuracy as TSV-M / \ftsv{}. \ftsv{} wins $29$ of the $30$ comparisons and loses only on GLUE at the coefficient of TSV-M, where its update is about half as large on the data; TSV-M run at the step of \ftsv{} stays within $0.56$ points of its own score in every setting.
\begin{table}
    \small
    \caption{Common-coefficient diagnostic}
    \label{tab:ccd}
    \resizebox{\textwidth}{!}{
        \begin{tabular}{@{}lccccc@{}}
        \toprule
        Setting ($\alpha$ TSV-M / \ftsv{}) & both at own $\alpha$ & both at $\alpha$ of TSV-M & both at $\alpha$ of \ftsv{} & TSV-M at the step of \ftsv{} & size ratio \\
        \midrule
        \model{ViT-B/32}, $T{=}8$ (0.9 / 1.6) & 92.05 / \textbf{96.02} & 92.05 / \textbf{92.94} & 81.54 / \textbf{96.02} & 92.04 & 0.553 \\
        \model{ViT-B/32}, $T{=}14$ (0.9 / 1.6) & 88.61 / \textbf{93.68} & 88.61 / \textbf{91.66} & 77.05 / \textbf{93.68} & 88.48 & 0.578 \\
        \model{ViT-B/32}, $T{=}20$ (0.8 / 1.5) & 85.91 / \textbf{91.83} & 85.91 / \textbf{89.41} & 74.71 / \textbf{91.83} & 85.64 & 0.573 \\
        \model{ViT-B/16}, $T{=}8$ (0.9 / 1.5) & 92.48 / \textbf{96.39} & 92.48 / \textbf{94.03} & 83.83 / \textbf{96.39} & 92.29 & 0.538 \\
        \model{ViT-B/16}, $T{=}14$ (0.9 / 1.7) & 89.31 / \textbf{93.60} & 89.31 / \textbf{92.24} & 75.50 / \textbf{93.60} & 88.81 & 0.583 \\
        \model{ViT-B/16}, $T{=}20$ (0.8 / 1.6) & 86.95 / \textbf{92.32} & 86.95 / \textbf{89.27} & 74.09 / \textbf{92.32} & 86.40 & 0.571 \\
        \model{ViT-L/14}, $T{=}8$ (1.1 / 2.2) & 96.67 / \textbf{98.87} & 96.67 / \textbf{97.83} & 87.70 / \textbf{98.87} & 96.67 & 0.499 \\
        \model{ViT-L/14}, $T{=}14$ (1.0 / 2.1) & 94.87 / \textbf{97.85} & 94.87 / \textbf{96.86} & 83.76 / \textbf{97.85} & 94.49 & 0.554 \\
        \model{ViT-L/14}, $T{=}20$ (1.0 / 1.8) & 92.97 / \textbf{97.68} & 92.97 / \textbf{96.07} & 85.29 / \textbf{97.68} & 93.05 & 0.533 \\
        \model{RoBERTa}, GLUE (1.4 / 3.0) & 78.01 / \textbf{87.86} & \textbf{78.01} / 66.69 & 74.03 / \textbf{87.86} & 78.57 & 0.491 \\
        \bottomrule
        \end{tabular}
    }
\end{table}

\textbf{Selection example:} with $k=1$, let two right singular directions $a,b$ of $\Delta_t$, also eigenvectors of $C_t$, change the output by $1.0$ and $0.6$ per unit input, with input standard deviations $0.3$ and $1.5$. Euclidean selection keeps $a$; data-aware selection compares $1.0\times0.3=0.30$ with $0.6\times1.5=0.90$, the singular values of $\Delta_tC_t^{1/2}$, and keeps $b$; in both cases the kept block is the expert's own action on the kept direction. 

\textbf{Orthogonalization example:} two tasks keep one direction each, at $10^\circ$ and $60^\circ$, and $\Cb$ has variance $1.8$ along $30^\circ$ and $0.2$ along $120^\circ$, so the outputs of the two directions correlate at $0.95$ on the pooled inputs. The Euclidean polar turns them to $-10^\circ$ and $80^\circ$, orthogonal in weight coordinates but still correlated at $0.80$ on the inputs; the metric polar $(\hat V^{\top}\Cb\hat V)^{-1/2}\hat V^{\top}$ makes them uncorrelated, $143^\circ$ apart, with Euclidean lengths $1.58$ and $1.75$.

\textbf{What data-aware selection changes.} Data-aware selection keeps directions of higher pooled input variance ($2.1$--$4.0$ against $1.4$--$1.8$ times the average on vision, $0.94$ against $0.71$ on GLUE), with Euclidean selection, $79\%$ of the kept directions fall below the average on GLUE and at most $6\%$ on vision, consistent with selection that is helping on GLUE.

\textbf{Summary.} \ftsv{} keeps the expert's own action on the subspace that the metric chooses, with its own singular values, and the metric never multiplies that content. The orthogonalization then keeps those singular values and those subspaces, and makes the retained directions orthonormal in the data metric instead of the Euclidean one. The merged update therefore carries the kept singular values $\hat s$ as its singular values in the data metric rather than in weight coordinates, and the resulting change of overall scale is absorbed by $\alpha$.

\section{Functional-maps dictionary}
\label{app:fmaps}
\begin{table}[b]
\centering
\small
\caption{The correspondence between functional maps and model merging for functional TSV.}
\label{tab:fmaps}
\begin{tabular}{@{}p{0.46\linewidth}p{0.46\linewidth}@{}}
\toprule
\textbf{Functional maps} \citep{ovsjanikov2012functional} &
\textbf{This manuscript} \\
\midrule
A correspondence is a linear map between \emph{function spaces}, not a
point-to-point map. & A task update is a linear operator between \emph{feature
spaces}, not a list of weights. \\
The map is written in a chosen basis with a geometry-aware (area-weighted)
inner product. & The operator is analyzed in the data-induced inner products
$\langle v,v'\rangle_C$ and $\langle u,u'\rangle_G$. \\
The basis is truncated to few low-frequency eigenfunctions. & The operator is
truncated to the top-$k$ directions of $G^{\,p_g}\Delta_tC^{\,p}$. \\
Maps from isometries are orthonormal \emph{in the function inner product}. &
The retained directions are made orthonormal in $\Cb$ and $\Gb$. \\
The map is recovered by least squares preserving descriptors. & The merge is measured by the functional error $\sum_t\tr[(\Delta^{\star}-\Delta_t)C_t(\Delta^{\star}-\Delta_t)^{\top}]$ of \eqref{eq:jls}. \\
The point map is recovered from the funct. map. & The merged weights are
recovered as $\theta_0+\alpha\Delta^{\star}$. \\
\bottomrule
\end{tabular}
\end{table}
Functional maps \citep{ovsjanikov2012functional,ovsjanikov2017course} represent a correspondence between two shapes by the linear operator it induces between their function spaces. That operator is written in truncated eigenbases that are orthonormal under each shape's area-weighted inner product $\langle f,g\rangle=f^{\top}Ag$, with $A$ the mass matrix of the mesh. Table~\ref{tab:fmaps} lists the counterpart of each ingredient in this paper. The task update $\Delta_t$ is the operator, weight coordinates play the role of mesh vertices, and the two mass matrices become the two metrics of a layer: $C$ on its inputs, with $\langle\Delta,\Delta'\rangle_C=\E_x\langle\Delta x,\Delta'x\rangle=\tr(\Delta C\Delta'^{\top})$, and $G$ on its outputs. Setting both metrics to the identity recovers TSV-M (Proposition~\ref{prop:nesting}); it corresponds to setting the mass matrices to the identity, so that updates are measured by their weight coordinates alone. At the derived powers $p=p_g=\tfrac12$, truncation (Step~1) and orthonormality (Step~2) are defined in these inner products, as they are for the eigenbasis of a functional map. The correspondence is structural: the eigenbasis of a functional map is fixed by the shape, whereas the directions retained here are computed per task from the whitened update.

\begin{table}[h]
\centering
\caption{\textbf{The token-embedding metric on MergeBench} (\model{Llama-3.2-3B}, five experts, all five pillars, $p{=}\tfrac12$, each row at its own validation-selected $\alpha$). The chain of Figure~\ref{fig:mechanism} plus \ftsvtwo{}, with the tied token embedding merged in the Euclidean metric and in the metric of \eqref{eq:embedmetric}. TSV-M carries no metric anywhere and is common to both blocks. Differences below ${\sim}0.2$ on the average are within the evaluation spread of Appendix~\ref{app:details}. \ftsv{} correspond to $\Delta_{\text{ort}}$.}
\label{tab:embed}
\small
\setlength{\tabcolsep}{5pt}
\begin{tabular}{@{}ll cccccc c@{}}
\toprule
Embedding metric & Row & Instr. & Math & Coding & Multiling. & Safety & \textbf{Avg.} & $\alpha$ \\
\midrule
  & TSV-M & 20.20 & 50.42 & 40.41 & 41.87 & 44.34 & 39.45 & 0.8 \\
\midrule
\multirow{3}{*}{Euclidean}
  & $\Delta_{\text{sel}}$ & 21.63 & 50.87 & 41.27 & 41.73 & 44.95 & 40.09 & 0.9 \\
  & \ftsv{} & 19.68 & 51.78 & 38.47 & 40.47 & 40.66 & 38.21 & 1.6 \\
  & \ftsvtwo{} & 18.05 & 51.93 & 36.12 & 40.50 & 40.15 & 37.35 & 1.5 \\
\midrule
\multirow{3}{*}{Eq.~\eqref{eq:embedmetric}}
  & $\Delta_{\text{sel}}$ & 18.81 & 49.05 & 39.60 & 41.80 & 43.72 & 38.60 & 0.8 \\
  & \ftsv{} & 21.14 & 57.01 & 36.46 & 40.30 & 44.10 & 39.80 & 1.8 \\
  & \ftsvtwo{} & 20.53 & 56.33 & 34.53 & 40.41 & 46.84 & 39.73 & 1.8 \\
\bottomrule
\end{tabular}
\end{table}
\section{Token-embedding metric of a causal language model}
\label{app:embed}

A causal LM merges one large matrix that no forward hook on a linear layer can observe: the token embedding $U\in\R^{|\mathcal V|\times n}$, tied to the output head, $128256\times3072$ on \model{Llama-3.2-3B}. It holds $12.3\%$ of the parameters and, measured on the five MergeBench experts, between $13.6\%$ (safety) and $66.0\%$ (multilingual) of each expert's task-vector energy $\norm{\tau_t}^2$ ($42.0\%$ math, $17.1\%$ coding, $14.2\%$ instruction).
Its input is nonetheless well defined. Read as the lookup $h_0=U^{\top}e_v$, with $e_v$ the one-hot token, a perturbation $\Delta$ of $U$ changes the final representation by $J\,\Delta^{\top}e_v$ to first order, and the functional norm of \eqref{eq:functionalreading} becomes
\begin{equation}
    \begin{split}
      \E_v\norm{J\,\Delta^{\top}e_v}^2
      =\tr\!\big(\Delta\,G_{h}\,\Delta^{\top}F\big)
      =\big\|F^{1/2}\,\Delta\,G_{h}^{1/2}\big\|_F^2, \\
      \quad F=\diag(\text{token frequency}),\quad G_h=\E[J^{\top}J],
    \end{split}
  \label{eq:embedmetric}
\end{equation}
so the token frequency is the metric on the vocabulary side and the downstream sensitivity at the embedding output, estimated with the same Gaussian VJP probes as $G_t$ in \S\ref{sec:ftsv2}, is the metric on the hidden side. In the stored $|\mathcal V|\times n$ layout, $G_h$ occupies the right factor of \eqref{eq:twosided} and $F$ the left one: \ftsv{}, which consumes the right factor only, applies the hidden-side sensitivity, and \ftsvtwo{} applies both. The vocabulary-side factor is diagonal, which is what makes the layer tractable: a dense one would be $61.3$\,GiB before any eigendecomposition, whereas the diagonal costs $0.5$\,MB, and every step of Appendix~\ref{app:conditioning} acts on eigenvalues only, so the machinery is unchanged. On this matrix the whole merge, five exact SVDs and both polar factors, takes $5.2$\,s at $15.3$\,GiB peak memory. The statistics are estimated from $128\times2$ held-out sequences per domain with $8$ probes each (Table~\ref{tab:budgets}). The metric is active in every MergeBench row of this paper and does not exist on the encoder pipelines.

Table~\ref{tab:embed} repeats the chain of Figure~\ref{fig:mechanism}, extended by \ftsvtwo{}, with the embedding merged in the Euclidean metric and in the metric of \eqref{eq:embedmetric}, on all five pillars. With the embedding outside the metric, metric-guided selection is worth $+0.64$ on the five-pillar average and the pooled orthogonalization $-1.87$; with it inside, the two increments are $-0.85$ and $+1.20$, the signs reported in Figure~\ref{fig:mechanism}. The output metric changes in the same way, from $-0.86$ against the one-sided method to $-0.07$. On a causal LLM, the pooled-metric orthogonalization therefore requires the token-embedding matrix to be inside the metric it orthogonalizes in.

\section{The choice of the spectral block}
\label{app:blockortho}

Table~\ref{tab:blockortho} shows the replacement of only the spectral block of \eqref{eq:select} in Step 1 with the exact minimizer $\tilde U^{(k)}\tilde U^{(k)\top}\Delta_t$, described in \S\ref{sec:ftsv}, with everything else fixed:  the conditioning, $p = 1
/2$, the rank budget and the coefficient selection. The results show that \eqref{eq:select} is the better option on average across all benchmarks under both orthogonalizations, and that the pooled-metric orthogonalization, improves on the Euclidean one with either spectral blocks.
\begin{table}[h]
\centering
\caption{\textbf{Spectral block $\times$ orthogonalization.} Normalized score: mean over the nine vision settings of Table~\ref{tab:vision}, GLUE, and the MergeBench five-pillar average. In parentheses: \eqref{eq:select} minus the exact block.}
\label{tab:blockortho}
\small
\begin{tabular}{@{}l cc@{}}
\toprule
& Euclidean orthogonalization & Pooled-metric orthogonalization (ours) \\
& ours eq. \ref{eq:select} / exact (gap) & ours eq. \ref{eq:select} / exact (gap) \\
\midrule
Vision (mean of nine settings) & \textbf{91.20} / 90.90 ($+0.30$) & \textbf{95.36} / 95.27 ($+0.09$) \\
GLUE & \textbf{80.42} / 75.66 ($+4.76$) & \textbf{87.86} / 84.95 ($+2.91$) \\
MergeBench & 38.60 / \textbf{39.14} ($-0.54$) & \textbf{39.80} / 39.32 ($+0.48$) \\
\bottomrule
\end{tabular}
\end{table}

\section{Sample-level stability of the data metric}
\label{app:stability}

The method reaches the data through one object per layer, the input second moment $C$, estimated from $N=512$ unlabeled examples per task. Everything downstream, which subspace each expert keeps and what cross-task orthogonality means, is a function of $C$, so the uncertainty that matters is over \emph{which} $512$ examples define it. We measure it first on the metric itself, without merging or evaluating, and then on the accuracy of the merge.
\begin{table}[b]
\centering
\caption{\textbf{Two draws of the same statistics sample} (\model{ViT-B/32}, $T{=}8$, $r{=}128$, $p{=}\tfrac12$). Median relative distance between two independent draws of $N{=}512$ examples over the $48$ observable matrices ($\times8$ tasks for per-task metrics), at the three stages of Appendix~\ref{app:conditioning}, with the interquartile range for the whitening factor. The last row is the distance between two different tasks' metrics from the same draw, the scale the others are read on.}
\label{tab:stability}
\small
\begin{tabular}{@{}lccc@{}}
\toprule
Two draws of $N{=}512$ examples & raw $C$ & $\cond(C;r)$ & $C^{\,p}$ \\
\midrule
per-task $C_t$ (1152 pairs) & 0.064 & 0.062 & 0.065 [0.051, 0.095] \\
pooled $\Cb$ (144 pairs) & 0.028 & 0.028 & 0.031 [0.024, 0.044] \\
different tasks, same draw (1344 pairs) & 1.139 & 1.112 & 0.547 [0.440, 0.636] \\
\bottomrule
\end{tabular}
\end{table}
\textbf{Setup.} On \model{ViT-B/32} with the $8$-task suite, at the setting of the reported results ($r=128$, $p=\tfrac12$, base encoder, training split), we collect three independent draws (seeds $0$, $1$, $2$) of every per-task statistic, together with draws of $N\in\{128,256,1024\}$, class-balanced draws, draws from the validation and test splits, draws through the finetuned experts, and two disjoint halves of one permutation: $27$ draws in all, compared layer by layer. The distance between two estimates $A,B$ of the same layer's metric is $\norm{A-B}_F/\operatorname{mean}(\norm{A}_F,\norm{B}_F)$ on unit-scaled metrics, taken at the three stages the merge puts $C$ through: raw, conditioned, and the whitening factor $C^{\,p}$ the merge multiplies by. The agreement of two kept top-$r$ eigenspaces is $\norm{U_A^{\top}U_B}_F^2/r$, reported as a multiple of its chance level $r/n$, which is $0.167$ at width $768$ and $0.042$ at width $3072$. Every number is a median over the $48$ observable matrices, times the $8$ tasks for per-task metrics, and is read against the distance between two \emph{different tasks'} metrics from the same draw: the scale on which a distance between two estimates of the same quantity is small or large.

\begin{figure}
\centering
\includegraphics[width=0.9\linewidth]{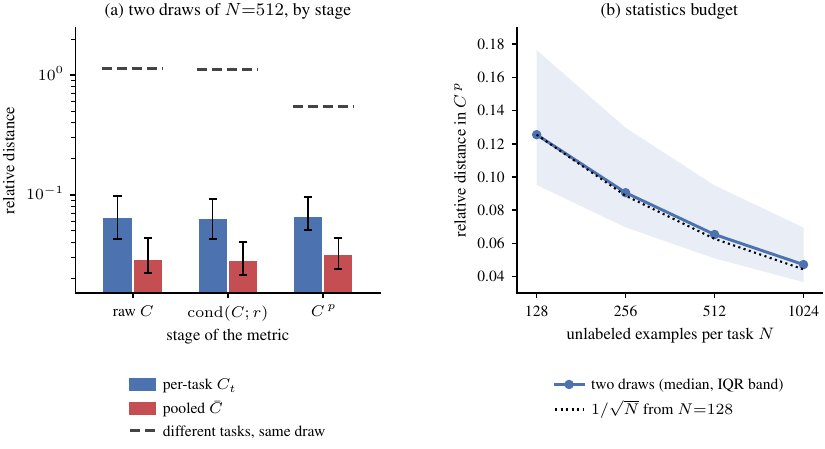}
\caption{\textbf{Sample-level stability of the input metric} (\model{ViT-B/32}, $T{=}8$). (a)~Median relative distance between two independent draws of $N{=}512$ examples at the three stages the merge puts $C$ through, for the per-task and the pooled metric (bars with interquartile whiskers, log scale); the dashed marks are the distance between two different tasks' metrics from the same draw. (b)~The same distance on $C^{\,p}$ against the budget $N$, with its interquartile band and the $1/\sqrt N$ law anchored at $N{=}128$.}
\label{fig:stability}
\end{figure}

\textbf{Redrawing the sample is a second-order effect.} Two draws of $512$ examples differ by $0.064$ on the raw metric and $0.065$ on $C^{\,p}$, against $1.139$ and $0.547$ between two different tasks: $17.7\times$ and $8.4\times$ less (Table~\ref{tab:stability}, Figure~\ref{fig:stability}a). The kept top-$r$ eigenspace of a redraw agrees at $5.6\times$ chance, against $3.3\times$ for two different tasks.

\textbf{Pooling, not conditioning, averages the sample out.} Conditioning leaves the redraw distance essentially unchanged ($0.064\to0.062$); its role is the bound on $\kappa$ of Appendix~\ref{app:conditioning}. Pooling over the $T=8$ tasks reduces the distance by $2.3\times$ ($0.028$ on the raw metric, $0.031$ on $\Cb^{\,p}$), close to the $\sqrt8$ that independent averaging would give. The pooled metric, which the orthogonalization step inverts and which Figure~\ref{fig:mechanism} identifies as the carrier of the gain, is thus the most stable object in the pipeline, $40\times$ below the cross-task distance it encodes, and its condition number changes by $0.35\%$ in the median and at most $4.1\%$ across draws.

\begin{table}[b]
\centering
\caption{\textbf{Other meanings of a different sample.} Median relative distance between two per-task estimates on the whitening factor $C^{\,p}$ (\model{ViT-B/32}, $T{=}8$), for each way in which the two estimates can differ; the different-task distance is the reference.}
\label{tab:stability_sources}
\small
\begin{tabular}{@{}lc@{}}
\toprule
What differs between the two estimates (per-task, on $C^{\,p}$) & distance \\
\midrule
which $512$ examples, random draws & 0.065 \\
which $512$ examples, class-balanced draws & 0.063 \\
random vs.\ class-balanced draw & 0.068 \\
training vs.\ validation vs.\ test split & 0.074 \\
base encoder vs.\ finetuned expert & 0.378 \\
different task (reference) & 0.547 \\
\bottomrule
\end{tabular}
\end{table}
\textbf{Other ways the sample can differ.} Table~\ref{tab:stability_sources} replaces the second random draw by other meanings of ``a different sample''. Class-balanced draws, the choice between random and balanced sampling, and the choice of split are indistinguishable from a plain redraw ($0.063$ to $0.074$ on $C^{\,p}$). The one choice that moves the metric materially is whose activations define it: through the finetuned expert instead of the base encoder the distance is $0.378$, close to the different-task level of $0.547$. Every reported metric is collected through the shared base.

\textbf{The estimator follows $1/\sqrt N$.} From $N=128$ to $1024$ the redraw distance on $C^{\,p}$ falls as $0.125$, $0.091$, $0.065$, $0.047$, within a few percent of the $1/\sqrt N$ law anchored at $N=128$ ($0.125$, $0.089$, $0.063$, $0.044$), and the kept subspace converges monotonically, from $4.99\times$ to $5.77\times$ chance (Table~\ref{tab:stability_budget}, Figure~\ref{fig:stability}b). Two disjoint halves of one permutation, an independence mechanism unrelated to reseeding, give $0.092$ at $N=256$ against $0.091$ across seeds, so the seed-to-seed spread is sampling noise and nothing else. Doubling the budget to $1024$ would reduce an already second-order quantity by $1.4\times$.

\begin{table}
\centering
\caption{\textbf{The statistics budget.} Distance between two draws on $C^{\,p}$ against the budget $N$ (\model{ViT-B/32}, $T{=}8$), the $1/\sqrt N$ law anchored at $N{=}128$, and the agreement of the kept top-$r$ eigenspace as a multiple of chance. At $N{=}256$, two disjoint halves of one permutation give $0.092$.}
\label{tab:stability_budget}
\small
\begin{tabular}{@{}lcccc@{}}
\toprule
Budget $N$ & 128 & 256 & 512 & 1024 \\
\midrule
two draws, on $C^{\,p}$ & 0.125 & 0.091 & 0.065 & 0.047 \\
$1/\sqrt N$ from $N{=}128$ & 0.125 & 0.089 & 0.063 & 0.044 \\
top-$r$ overlap ($\times$ chance) & 4.99 & 5.36 & 5.60 & 5.77 \\
\bottomrule
\end{tabular}
\end{table}

\textbf{Accuracy over draws.} Table~\ref{tab:stability_acc} measures how much the merged model itself moves: five independent draws (seeds $0$ to $4$) of the $N=512$ examples per task, each merged and evaluated with its own validation-selected $\alpha$, on \model{ViT-B/32} ($T{=}8$) and on GLUE. The standard deviation of the normalized accuracy of \ftsv{} is $0.11$ on vision and $0.48$ on GLUE, and the selected $\alpha$ varies by at most $0.2$ across draws. The mean increment of the pooled orthogonalization over selection alone, $3.76$ and $7.97$ normalized points, exceeds the standard deviation of either row by more than an order of magnitude.

The metric-level study covers one backbone, one task count and the input metric only; the accuracy spread is measured on \model{ViT-B/32} ($T{=}8$) and on GLUE.
\begin{table}[h]
\centering
\caption{\textbf{Accuracy over draws of the statistics sample.} Normalized test accuracy over five independent draws (seeds $0$ to $4$) of the $N{=}512$ examples per task, mean $\pm$ standard deviation, each draw merged and evaluated with its own validation-selected $\alpha$, whose range is given below it. $\Delta_{\text{sel}}$ is metric-guided selection without the pooled orthogonalization, as in Table~\ref{tab:alphas}.}
\label{tab:stability_acc}
\small
\begin{tabular}{llccc}
\toprule
& & $\Delta_{\text{sel}}$ & \ftsv{} & \ftsvtwo{} \\
\midrule
\multirow{2}{*}{\model{ViT-B/32}, $T{=}8$}
  & normalized, mean $\pm$ sd & $92.09 \pm 0.12$ & $95.85 \pm 0.11$ & $95.97 \pm 0.07$ \\
  & selected $\alpha$ range    & $1.0$--$1.1$ & $1.6$--$1.7$ & $1.6$--$1.8$ \\
\cdashlinelr{1-5}
\multirow{2}{*}{RoBERTa (GLUE)}
  & normalized, mean $\pm$ sd & $79.89 \pm 0.17$ & $87.86 \pm 0.48$ & $87.91 \pm 0.31$ \\
  & selected $\alpha$ range    & $1.4$--$1.5$ & $3.0$--$3.2$ & $3.2$--$3.4$ \\
\bottomrule
\end{tabular}
\end{table}

\section{The whitening power: full grid}
\label{app:power}

Table~\ref{tab:power_full} extends the grid of \S\ref{sec:power} to every power that was run, with the coefficient each cell selected. All cells of a row share the cached statistics: the power is applied to the conditioned metric at merge time, not at collection time. On GLUE the cells at $p\notin\{0,\tfrac12\}$ were swept on the coarser grid $\{0.5,1.0,\dots,6.0\}$; the $p=\tfrac12$ cell selects $\alpha=3.0$ on both grids and is the run of Table~\ref{tab:glue}. Choosing $p$ by the validation objective instead of by derivation returns $\tfrac12$ on every modality: the validation optimum over $\alpha$ is highest at $p=\tfrac12$ on \model{ViT-B/32} and on GLUE, and the validation likelihood is lowest there on MergeBench. The collapse at large $p$ is the orthogonalization's inverse factor at work: selection never multiplies kept content by a power of $C$, whereas \eqref{eq:metricortho} amplifies by up to $\kappa(\Cb)^{p}$, which on \model{ViT-B/32}'s worst-conditioned layer is $38$ at $p=\tfrac12$ and ${\sim}10^{12}$ at $p=4$. The PAVE-style column is \ftsv{} at $p=1$, the power at which PAVE ranks directions, with our projection and cross-task orthogonalization, PAVE itself is compared in Appendix~\ref{app:newbaselines}.

\paragraph{The whitening power extended results.}
To clarify the role of the whitening power, we provide additional results here to clarify \S\ref{sec:power} in the main manuscript.
The power enters twice, and the derivation constrains only one of the two: selection uses $C_t^{\,p}$ to rank directions, which any $p>0$ does, whereas the orthogonalization's constraint $R\,\Cb^{\,2p}R^{\top}=I$ coincides with $\Cb$-orthonormality only at $p=\tfrac12$ (Proposition~\ref{prop:exponent}). Sweeping the two independently over $\{0,\tfrac14,\tfrac12,1,2\}^2$ separates them: on both \model{ViT-B/32} ($T{=}8$) and GLUE the orthogonalization power $\tfrac12$ is optimal in \emph{every} row of the grid, for every selection power, while the selection power is comparatively flat ($96.13$ at $p_{\text{sel}}{=}\tfrac14$ against $96.02$ at $\tfrac12$ on vision, inside the $0.11$ standard deviation of the draw). The quantitative prediction the derivation makes is therefore confirmed exactly where it is made, and the earlier single-$p$ sweep was measuring the orthogonalization.

Refining the grid to $\{0.3,0.4,0.5,0.6,0.7\}$ places the vision optimum at $p{=}0.4$ ($96.07$) against $96.02$ at $\tfrac12$, a separation of $0.05$ points not distinguishable inside the $0.11$ standard deviation of the statistics draw. On GLUE the optimum is $\tfrac12$ ($87.86$, against $87.68$ at $0.4$). We therefore claim that the derived exponent is not improved upon, rather than that it is uniquely optimal.

\begin{table}[t]
\centering
\caption{\textbf{The whitening power, full grid.} Every cell is one run at its own validation-selected $\alpha$ (second line of each block), on the conditioning recipe of Appendix~\ref{app:conditioning}. Vision and GLUE report absolute accuracy with normalized in parentheses; MergeBench the five-pillar average and the teacher-forced validation likelihood at the selected $\alpha$. The $p{=}0$ column is the metric switched off: the $c_{\text{sites}}{=}\texttt{none}$ control on vision and GLUE, and \ftsv{} run at $p{=}0$ on MergeBench, which Proposition~\ref{prop:nesting} makes TSV-M and which is a separate evaluation from the TSV-M row of Table~\ref{tab:mergebench}. $^{\dagger}$: the validation optimum sat on the grid floor ($\alpha{=}0.1$), so the cell bounds the power rather than measuring it.}
\label{tab:power_full}
\small
\setlength{\tabcolsep}{4pt}
\resizebox{\textwidth}{!}{
\begin{tabular}{@{}lccccc@{}}
\toprule
$p$ & $0$ (TSV-M) & $0.25$ & $0.5$ (derived) & $1.0$ (PAVE-style) & $2.0$ \\
\midrule
\model{ViT-B/32} $T{=}8$ & 83.41 (92.05) & 86.47 (95.52) & \textbf{86.81 (96.02)} & 85.92 (95.07) & 84.93 (93.94) \\
\quad selected $\alpha$ & 0.9 & 1.5 & 1.6 & 1.6 & 1.3 \\
\midrule
\model{RoBERTa} GLUE & 66.35 (78.01) & 71.28 (85.29) & \textbf{73.13 (87.86)} & 68.40 (82.52) & 56.60 (68.16) \\
\quad selected $\alpha$ & 1.4 & 2.0 & 3.0 & 3.0 & 1.5 \\
\midrule
\model{Llama-3.2-3B} MergeBench & 39.37 & 39.67 & \textbf{39.76} & 36.46 & 13.85$^{\dagger}$ \\
\quad selected $\alpha$ & 0.8 & 1.3 & 1.8 & 1.7 & 0.1$^{\dagger}$ \\
\quad validation NLL & 1.6869 & 1.6415 & 1.6404 & 1.6666 & 8.5171 \\
\bottomrule
\end{tabular}
}
\end{table}

\section{Site attribution: definitions and the GLUE panels}
\label{app:sites}

\textbf{Sites and groups.} A site is a weight matrix whose input (for $C$) or output (for $G$) is observable by a forward hook on the base encoder. On the CLIP encoders these are, per block, the fused attention in-projection (q/k/v), the attention out-projection and the two MLP linears, $48$ matrices on \model{ViT-B/32} ($12$ blocks); on \model{RoBERTa-base} the query, key, value and attention-output projections and the two feed-forward linears, $72$ matrices; on \model{Llama-3.2-3B} the q, k, v and o projections and the gate, up and down projections, $196$ matrices ($28$ layers). The groups of Tables~\ref{tab:sites} and \ref{tab:sites_glue} are: \method{q/k/v}, the attention projections that do not write into the residual stream; \method{all attention}, those plus the attention out-projection; \method{MLP linears}, every non-attention site; \method{residual}, the two matrices per block whose output is added to the residual stream (attention out-projection and MLP down-projection); \method{non-residual}, its complement; and unions of these. A site outside the kept group receives the identity metric on that side and, by Proposition~\ref{prop:nesting}, is merged there exactly as by the method one rung down. Matrices that are not sites (token and positional embeddings, the convolutional stem, projection heads) take the Euclidean path; on \model{Llama-3.2-3B} the token embedding is the exception treated in Appendix~\ref{app:embed}. Table~\ref{tab:sites_glue} is the GLUE counterpart of Table~\ref{tab:sites}, discussed in \S\ref{sec:sites}.

\renewcommand{\arraystretch}{1.4}
\begin{table}[t]
    \centering
    \resizebox{1\columnwidth}{!}{
        \begin{tabular}{cl cc @{\hskip 1.6em} cl cc}
            \toprule
            & \multirow{2}{*}{\textbf{Sites kept}} & \multicolumn{2}{c}{\model{Input metric $C$}} & & \multirow{2}{*}{\textbf{Sites kept}} & \multicolumn{2}{c}{\model{Output metric $G$}} \\
            & & \multicolumn{2}{c}{\small RoBERTa, GLUE} & & & \multicolumn{2}{c}{\small RoBERTa, GLUE} \\
            \cmidrule(lr){2-2}\cmidrule(lr){3-4} \cmidrule(lr){6-6}\cmidrule(lr){7-8}
            & & Score & Gain & & & Score & Gain \\
            \cmidrule(r){1-4} \cmidrule(l){6-8}
            \rowcolor{gray!15}
            \cellcolor{white} & \method{none ($\equiv$\,TSV-M)} & $66.35_{(78.01)}$ & --- & \cellcolor{white} & \method{none ($\equiv$\,\ftsv{})} & $73.13_{(87.86)}$ & --- \\
            & \method{q/k/v} & $69.84_{(82.79)}$ & $+4.78$ &  & \method{q/k/v} & $72.54_{(87.41)}$ & $-0.45$ \\
            & \method{all attention} & $70.45_{(83.73)}$ & $+5.72$ &  & \method{all attention} & $72.80_{(87.70)}$ & $-0.16$ \\
            & \method{non-residual} & $71.41_{(84.54)}$ & $+6.53$ &  & \method{non-residual} & $72.99_{(87.84)}$ & $-0.02$ \\
            & \method{residual} & $68.20_{(80.94)}$ & $+2.93$ &  & \method{residual} & $73.39_{(88.30)}$ & $+0.44$ \\
            & \method{q/k/v $+$ residual} & $70.65_{(85.14)}$ & $+7.13$ &  & \method{q/k/v $+$ residual} & $73.05_{(87.87)}$ & $+0.01$ \\
            & \method{MLP linears} & $69.72_{(82.58)}$ & $+4.57$ &  & \method{MLP linears} & $73.96_{(88.95)}$ & $+1.09$ \\
            \rowcolor{mygreen!50}
            \cellcolor{white} & \method{MLP $+$ residual} & $69.99_{(83.35)}$ & $+5.33$ & \cellcolor{white} & \method{MLP $+$ residual} & $73.83_{(88.89)}$ & $+1.03$ \\
            \rowcolor{mygreen!50}
            \cellcolor{white} & \method{all sites} & $\mathbf{73.13_{(87.86)}}$ & $+9.85$ & \cellcolor{white} & \method{all sites} & $\mathbf{73.89_{(88.98)}}$ & $+1.12$ \\
            \bottomrule
        \end{tabular}
    }
    \caption{\textbf{Site attribution on GLUE: the same ablation as Table~\ref{tab:sites}, on the other modality.} RoBERTa-base, 8 GLUE tasks, both panels; which weight matrices keep their measured metric, with dropped sites behaving exactly as the method one rung down (Proposition~\ref{prop:nesting}). Absolute score with normalized in subscript; \emph{Gain} is the normalized-score change against \method{none}. The output panel uses the Gaussian probes on the encoder's final representation, as every two-sided result in this paper (Appendix~\ref{app:details}).}
    \label{tab:sites_glue}
\end{table}

\section{Experimental protocol and reproduction}
\label{app:details}

\textbf{Backbones, experts and task suites.} \emph{Vision:} OpenCLIP ViT-B/32, ViT-B/16 and ViT-L/14 image encoders with the public per-task experts used by MASS \citep{mass2025}. The $8$-task suite \citep{ilharco2023task} is SUN397 \citep{xiao_sun_2016}, Cars \citep{krause20133d}, RESISC45 \citep{cheng_remote_2017}, EuroSAT \citep{helber_eurosat_2019}, SVHN \citep{netzer_reading_nodate}, GTSRB \citep{stallkamp_german_2011}, MNIST \citep{noauthor_mnist_nodate} and DTD \citep{cimpoi_describing_2014}; the $14$-task suite \citep{wang2024tall} adds Flowers102 \citep{nilsback_automated_2008}, PCAM \citep{veeling_rotation_2018}, FER2013 \citep{goodfellow_challenges_2013}, OxfordIIITPet \citep{parkhi_cats_2012}, STL10 \citep{coates_analysis_2011} and CIFAR100 \citep{krizhevsky_learning_nodate}; the $20$-task suite adds CIFAR10 \citep{krizhevsky_learning_nodate}, Food101 \citep{bossard_food-101_2014}, FashionMNIST \citep{xiao_fashion-mnist_2017}, EMNIST \citep{cohen_emnist_2017}, KMNIST \citep{clanuwat_deep_2018} and RenderedSST2 \citep{socher_recursive_nodate}. Each task is evaluated with its frozen zero-shot CLIP head. \emph{Language:} \texttt{roberta-base} with the RoBERTa-base GLUE experts distributed with WUDI-Merging \citep{cheng2025wudi} and reused by ESM \citep{esm2026}, on CoLA, SST-2, MRPC, STS-B, QQP, MNLI, QNLI and RTE \citep{wang2018glue}; each expert's classification head is split from its checkpoint and swapped in at evaluation, never merged. 
Every baseline row of Tables~\ref{tab:vision} and \ref{tab:glue} was produced by the same evaluation code and from the same experts downloaded from Huggingface, under the coefficient protocol above, and each cell is a single run whose configuration is stored beside its score. The vision experts are the publicly accessible per-task checkpoints on Huggingface released by \citet{mass2025}. Because the normalized score is divided by the same reference for every method, that offset cancels in the comparison that the tables make.
\emph{Generative:} \texttt{meta-llama/Llama-3.2-3B} in bf16 with the five MergeBench experts \texttt{Llama-3.2-3B\_\{instruction, math, coding, multilingual, safety\}} \citep{mergebench2025}, the embedding and output are the same one matrix and are merged once. 

\begin{table}
\centering
\caption{Statistics budgets and defaults of every reported cell. Examples are counted per task; a batch of $G$ carries its $n_p$ probes. Ranks on the encoders are absolute; on \model{Llama-3.2-3B} they are fractions of each layer's width, resolved per layer.}
\label{tab:budgets}
\small
    \resizebox{1\textwidth}{!}{
        \begin{tabular}{@{}llll@{}}
        \toprule
         & & Vision (CLIP), GLUE (RoBERTa) & MergeBench (\model{Llama-3.2-3B}) \\
        \midrule
        \multirow{2}{*}{input metric $C$} & examples per task & $4\times128=512$ & $63\times16=1008$ sequences ($\le4096$ tokens) \\
         & rank $r$ & $128$ & $0.2\times$ width ($614$ of $3072$; $1638$ of $8192$) \\
        \midrule
        \multirow{3}{*}{output metric $G$} & examples per task & $8\times64=512$ & $128\times2=256$ sequences \\
         & probes per batch $n_p$ & $32$ & $8$ \\
         & rank $r_g$ & $256$ & $0.4\times$ width \\
        \midrule
        token-embedding metric & & --- & $128\times2$ sequences, $8$ probes \\
        whitening powers & $p=p_g$ & $\tfrac12$ & $\tfrac12$ \\
        activations & & base encoder, training split & base model, held-out sets \\
        \bottomrule
        \end{tabular}
    }
\end{table}
\textbf{Splits and scores.} On vision and GLUE the validation split is a seeded $10\%$ of each task's training set, the test split is the standard test set (for GLUE, the original validation set, MNLI uses the matched set), and the statistics are drawn from the remaining training data. Scores are accuracy, except Matthews correlation for CoLA and Spearman correlation for STS-B, floored at $0$ as in the GLUE tables. The normalized score divides each task's merged score by its expert's score and averages over tasks. On MergeBench, the statistics and the coefficient sweep both use the benchmark's per-domain held-out sets ($1000$ rows per domain under the alpaca SFT template), and the five pillars are scored on the benchmark's test tasks as described.

\textbf{Statistics budgets and defaults} are in Table~\ref{tab:budgets}. All statistics are collected through the base model, in one forward pass per batch for $C$ and one forward plus $n_p$ backward passes per batch for $G$, over token rows with padding masked; the Gaussian probes are drawn on the encoder's final representation (the CLIP image embedding; the final hidden states of the RoBERTa body and of the Llama decoder). The vision and GLUE rank budgets are absolute; the \model{Llama-3.2-3B} ones are fractions of each layer's width (Appendix~\ref{app:conditioning}). The rank kept per task and layer is $k=\max(1,\lfloor\min(m,n)/T\rfloor)$, so the $T$ stacked blocks fill the layer's spectrum exactly once.

\textbf{Coefficient selection.} The candidate grid is a property of the benchmark: $\{0.1,0.2,\dots,2.5\}$ on vision, $\{0.1,\dots,4.0\}$ on GLUE and $\{0.1,\dots,2.0\}$ on MergeBench, never including $0$, which is the pretrained model rather than a merge. The vision ceiling is set by \model{ViT-L/14}, whose optima reach $2.2$, and the GLUE ceiling by \method{Iso-CTS} and \ftsvtwo{}, whose optima exceed $3.0$. For vision and GLUE, each candidate is evaluated on the validation split, and the one with the highest average normalized score is retained. On MergeBench, a per-$\alpha$ generative evaluation is too expensive, so each candidate is scored by the mean teacher-forced negative log-likelihood of the held-out rows (prompt and response tokens, up to $1024$ tokens, equal-weighted over the five domains), and the lowest-likelihood $\alpha$ receives the single full evaluation. Every results file flags an optimum that sits on a grid boundary. One of the results reported here is flagged, at the grid's smallest coefficient $\alpha=0.1$: the $p{=}2.0$ MergeBench cells of Table~\ref{tab:power_full}. Table~\ref{tab:alphas} lists the selected coefficient of every cell of Tables~\ref{tab:vision}, \ref{tab:glue}, and \ref{tab:mergebench} produced by our methods and their in-family controls, none of which sit on a boundary.

\begin{table}
\centering
\caption{Selected coefficients $\alpha$ of the reported cells (Tables~\ref{tab:vision}, \ref{tab:glue}, and \ref{tab:mergebench} and Figure~\ref{fig:mechanism}). TSV-M is the $c_{\text{sites}}{=}\texttt{none}$ control and $\Delta_{\text{sel}}$ the metric-guided selection without the pooled orthogonalization. Grids: $0.1$ to $2.5$ on vision, to $4.0$ on GLUE, to $2.0$ on MergeBench.}
\label{tab:alphas}
\small
\begin{tabular}{@{}l ccc ccc ccc c c@{}}
\toprule
& \multicolumn{3}{c}{\model{ViT-B/32}} & \multicolumn{3}{c}{\model{ViT-B/16}} & \multicolumn{3}{c}{\model{ViT-L/14}} & GLUE & MergeBench \\
\cmidrule(lr){2-4}\cmidrule(lr){5-7}\cmidrule(lr){8-10}\cmidrule(lr){11-11}\cmidrule(lr){12-12}
$T$ & 8 & 14 & 20 & 8 & 14 & 20 & 8 & 14 & 20 & 8 & 5 \\
\midrule
TSV-M & 0.9 & 0.9 & 0.8 & 0.9 & 0.9 & 0.8 & 1.1 & 1.0 & 1.0 & 1.4 & 0.8 \\
$\Delta_{\text{sel}}$ & 1.1 & 1.0 & 0.9 & 1.0 & 1.0 & 0.9 & 1.3 & 1.1 & 1.1 & 1.4 & 0.8 \\
\ftsv{} & 1.6 & 1.6 & 1.5 & 1.5 & 1.7 & 1.6 & 2.2 & 2.1 & 1.8 & 3.0 & 1.8 \\
\ftsvtwo{} & 1.6 & 1.8 & 1.6 & 1.7 & 1.7 & 1.5 & 1.9 & 1.9 & 1.9 & 3.2 & 1.8 \\
\bottomrule
\end{tabular}
\end{table}

\textbf{Validation curves on MergeBench.} Table~\ref{tab:nllcurve} gives the likelihood that selects $\alpha$, both optima are interior to the grid, and on \model{Gemma-2-2B} the minima are $1.722$ at $\alpha=0.5$ (TSV-M) and $1.632$ at $\alpha=1.4$ (\ftsv{}).

\begin{table}[h]
\centering
\caption{\textbf{Validation curves on MergeBench} (\model{Llama-3.2-3B}): teacher-forced NLL of the held-out rows at each $\alpha$, the score that selects the coefficient of Table~\ref{tab:mergebench} (lower is better).}
\label{tab:nllcurve}
\small
    \begin{tabular}{@{}l ccccccccc@{}}
    \toprule
    $\alpha$ & 0.4 & 0.6 & 0.8 & 1.0 & 1.2 & 1.4 & 1.6 & 1.8 & 2.0 \\
    \midrule
    TSV-M & 1.717 & 1.691 & \textbf{1.688} & 1.696 & 1.719 & 1.755 & 1.807 & 1.876 & 1.961 \\
    \ftsv{} & 1.747 & 1.718 & 1.692 & 1.673 & 1.656 & 1.647 & 1.641 & \textbf{1.640} & 1.647 \\
    \bottomrule
    \end{tabular}
\end{table}

\textbf{Observability.} Statistics come from forward hooks on the base encoder. On the CLIP encoders the fused attention module is temporarily replaced by an unfused equivalent at collection time, so that both attention projections are observable on both sides. Matrices without observed statistics take the Euclidean path per Proposition~\ref{prop:nesting}; non-matrix parameters (biases, normalization gains) are averaged across tasks.

\textbf{Determinism, and the MergeBench spread.} Given fixed experts, cached statistics and the fixed validation split the merge is deterministic: three different runs return identical scores to every stored digit, and the vision and GLUE evaluations, being deterministic passes conducted on fixed splits, they inherit that determinism, so we quote no confidence intervals there. The MergeBench evaluation does not: two of its five pillars are scored by sampled generation. One configuration (\ftsv{} at $p=\tfrac12$ with the embedding metric, $\alpha=1.8$) evaluated three times returned five-pillar averages of $39.67$, $39.76$ and $39.80$: coding and multilingual were identical across the three, instruction and math moved by $0.21$ and $0.15$, and safety by $0.68$. We treat differences below about $0.2$ five-pillar points as not separating two MergeBench configurations. This covers the gap between \ftsv{} and \ftsvtwo{} in Table~\ref{tab:mergebench}, and the gap between that table's \ftsv{} row and the $p=\tfrac12$ cell of Table~\ref{tab:power_full}.

\textbf{MergeBench evaluation.} \emph{Instruction:} IFEval, the mean of prompt-level and instruction-level strict accuracy. \emph{Math:} GSM8K with $8$-shot chain of thought, flexible-extract exact match. \emph{Multilingual:} the twelve Okapi tasks (M-MMLU, M-ARC and M-HellaSwag in French, Spanish, German and Russian), accuracy, zero-shot. These three run in lm-eval-harness with a $4096$-token context and an automatically sized batch. \emph{Coding:} HumanEval+ and MBPP+ through the bigcode evaluation harness, pass@1 over $10$ samples at temperature $0.2$, top-$p$ $0.95$, $512$ new tokens, bf16, averaged over the two suites. \emph{Safety:} WildGuardTest (harm rate), HarmBench and DoAnythingNow (attack success rate) and XSTest (accuracy on its safe/unsafe contrast set) through safety-eval-fork with the Llama-3 input template; the first three are inverted so that higher is safer and the four are averaged. Base model, task list, few-shot counts, sampling configuration and safety flags are those of MergeBench's own evaluation script, and every run persists every metric key lm-eval reports, so alternative readings (normalized accuracy on M-ARC and M-HellaSwag, prompt-level-only IFEval) can be produced without a re-run. 

\textbf{Cost.} Everything is training-free. The one-time per-task cost is one forward pass over the statistics sample for $C$ and, two-sided, one forward plus $n_p$ backward passes per batch for $G$; statistics are cached and shared by every configuration, so an ablation cell costs one merge plus one evaluation. Relative to TSV-M, merging adds per layer one eigendecomposition per active metric and two thin factorizations (a QR and a second SVD of the kept block). On one RTX PRO 6000 Blackwell, collecting $C$ for the eight vision tasks takes $21$\,s on \model{ViT-B/32} and $54$\,s on \model{ViT-L/14}; on \model{Llama-3.2-3B} the token-embedding matrix is merged in $5.2$ seconds using $15.3$\,GiB (Appendix~\ref{app:embed}). The merged model is a single static checkpoint in the base architecture, with no routing, no extra parameters and no inference-time overhead.

\section{Related works focus: MaTS and PAVE}
\label{app:newbaselines}

\textbf{MaTS} \citep{tam2024mats} follow the recipe prescribed for fully finetuned models \citep{tam2024mats}: RegMean's objective per linear layer, solved by at most $100$ conjugate-gradient steps started from the Task Arithmetic merge, with the reference code's stopping rule, on each expert's input Gram matrices from $1024$ unlabeled training examples per task. In this recipe RegMean, MaTS and \ftsv{} share the curvature form $C_t\otimes I$ but take $C_t$ from different models, so on GLUE each method therefore also runs on the others' $C_t$ (Table~\ref{tab:matched}): each is strongest on its own statistics, but the setting of \ftsv{} needs half the examples and only the pretrained encoder.

\begin{table}
\centering
\caption{\textbf{GLUE under matched statistics.} GLUE average, absolute with normalized in subscript, when all methods in a row read the same $C_t$; no statistic uses labels. Collection: all eight tasks on one GPU, mean $\pm$ standard deviation over ten runs.}
\label{tab:matched}
\small
\resizebox{\textwidth}{!}{
\begin{tabular}{@{}lccc ccc@{}}
\toprule
$C_t$ computed from & Examples/task & Models run & Collection & \ftsv{} & RegMean & MaTS \\
\midrule
pretrained encoder (\ftsv{} style) & 512 & 1 & $8.1\pm0.2$\,s & $\mathbf{73.1_{(87.9)}}$ & $57.3_{(66.2)}$ & $57.0_{(65.9)}$ \\
each expert on its own task (RegMean style) & 1024 & 8 & $10.5\pm0.2$\,s & $63.2_{(76.5)}$ & $71.6_{(85.5)}$ & $\mathbf{71.7_{(85.5)}}$ \\
\bottomrule
\end{tabular}
}
\end{table}

\textbf{PAVE}  We reproduced PAVE \citep{an2025pave} from the authors' released code, including its rank allocation (their Algorithm~1), and verified collection, rank search and purification against that code on identical inputs (agreement to $10^{-5}$). \emph{Aggregation rule:} PAVE purifies each task vector and leaves the merge to a standard method; we pass its purified vectors, unchanged, to Task Arithmetic (TA) and to TSV-M, with $512$ training examples per task through each expert and $\rho$ and $\alpha$ selected on validation (Table~\ref{tab:pave}). Purified experts keep $99.2$ to $102.2\%$ of their accuracy alone, yet validation selects the mildest $\rho=63/64$ in every cell and PAVE stays below its host method, hinting that the gain of \ftsv{} comes from the cross-task step that PAVE lacks.

\begin{table}[h]
\centering
\caption{\textbf{PAVE.} Absolute accuracy, normalized in parentheses; TSV-M is the standalone host.}
\label{tab:pave}
\small
\begin{tabular}{@{}l ccccc@{}}
\toprule
& TA & PAVE $+$ TA & TSV-M & PAVE $+$ TSV-M & \ftsv{} \\
\midrule
\model{ViT-B/32}, $T{=}8$ & 70.4 (77.9) & 69.8 (77.2) & 83.4 (92.0) & 81.5 (89.9) & \textbf{86.8 (96.0)} \\
\model{ViT-B/32}, $T{=}14$ & 66.8 (75.2) & 66.7 (75.0) & 79.1 (88.6) & 76.9 (86.2) & \textbf{83.6 (93.7)} \\
\model{ViT-B/32}, $T{=}20$ & 64.0 (71.9) & 63.3 (71.1) & 76.8 (85.9) & 74.2 (83.1) & \textbf{82.2 (91.8)} \\
RoBERTa, GLUE & 57.1 (66.1) & 56.9 (66.0) & 66.3 (78.0) & 58.4 (68.2) & \textbf{73.1 (87.9)} \\
\bottomrule
\end{tabular}
\end{table}

\section{MergeBench on Gemma-2-2B}
\label{app:gemma}

Table~\ref{tab:gemma} merges the five MergeBench experts of \model{Gemma-2-2B} \citep{mergebench2025} with the setup already utilized for \model{Llama-3.2-3B} in Table~\ref{tab:mergebench} with the same code, budgets and protocol, $p=\tfrac12$, every row at its validation-selected $\alpha$ and scored on all five pillars. \ftsv{} and \ftsvtwo{} lead TSV-M by $3.87$ and $4.76$ points, and coding, which drops on \model{Llama-3.2-3B}, is stable here, so the coding loss is model-dependent, on \model{Gemma-2-2B} instruction and safety also rise.

\begin{table}[h]
    \centering
    \caption{\textbf{\model{Gemma-2-2B} on MergeBench}, each row at its validation-selected $\alpha$.}
        \begin{tabular}{cl cccccc}
            \toprule
            & \multirow{2}{*}{\textbf{Method}} & \multicolumn{6}{c}{\model{Gemma-2-2B} --- MergeBench} \\
            \cmidrule(lr){3-8}
            & & Instruct. & Math & Coding & Multiling. & Safety & \textbf{Avg.} \\
            \cmidrule(r){2-2} \cmidrule(lr){3-3}\cmidrule(lr){4-4}\cmidrule(lr){5-5}\cmidrule(lr){6-6}\cmidrule(lr){7-7}\cmidrule(lr){8-8} \rowcolor{gray!15}
            \cellcolor{white} & \method{Pre-trained} & $19.39$ & $27.29$ & $24.96$ & $38.23$ & $42.67$ & $30.51$ \\
            \rowcolor{gray!15} \cellcolor{white} \multirow{-2}{*}{\small \vertical{Base}} & \method{Finetuned} & $44.42$ & $56.41$ & $35.76$ & $40.68$ & $81.82$ & $51.82$ \\
            \cdashlinelr{1-8}
            \multirow{3}{*}{\small \vertical{Fixed}} & \method{Model Soup} & $22.80$ & $39.12$ & $30.87$ & $\mathbf{42.18}$ & $41.99$ & $35.39$ \\
            & \method{Task Arithmetic} & $22.76$ & $38.89$ & $30.33$ & $42.14$ & $41.86$ & $35.20$ \\
            & \method{TSV-M} & $22.61$ & $38.82$ & $31.28$ & $41.85$ & $44.25$ & $35.76$ \\
            \cdashlinelr{1-8}
            & \method{ESM} & $18.16$ & $33.66$ & $27.09$ & $39.73$ & $42.88$ & $32.31$ \\
            \rowcolor{mygreen!50} \cellcolor{white} & \method{\textbf{Func. TSV}} & $\mathbf{29.60}$ & $\mathbf{42.84}$ & $31.32$ & $40.50$ & $53.89$ & $39.63$ \\
            \rowcolor{mygreen!50} \cellcolor{white} \multirow{-3}{*}{\small \vertical{Data}} & \method{\textbf{Func. TSV}}$^{2}$ & $28.48$ & $42.53$ & $\mathbf{31.40}$ & $39.98$ & $\mathbf{60.21}$ & $\mathbf{40.52}$ \\
            \bottomrule
        \end{tabular}
    \label{tab:gemma}
\end{table}

\end{document}